\documentclass[10pt,journal,compsoc]{IEEEtran}
\ifCLASSOPTIONcompsoc
  \usepackage[nocompress]{cite}
\else
  \usepackage{cite}
\fi
\ifCLASSINFOpdf
\else
\fi

\usepackage{times}
\usepackage{epsfig}
\usepackage{graphicx}
\usepackage{amsmath}
\usepackage{amssymb}
\usepackage{color}
\usepackage{xcolor}
\usepackage{hyperref}
\usepackage{booktabs}
\usepackage{multirow}
\usepackage{xspace}

\newcommand{\etal}{\emph{et al.}\xspace}
\newcommand{\etc}{\emph{etc.}\xspace}
\newcommand{\eg}{\emph{e.g.}\xspace}
\newcommand{\ie}{\emph{i.e.}\xspace}
\newcommand{\vs}{\emph{vs.}\xspace}

\newcommand{\figLabel}{Figure\xspace}
\newcommand{\eqLabel}{Equation\xspace}
\newcommand{\secLabel}{Section\xspace}
\newcommand{\tblLabel}{Table\xspace}
\newcommand{\mysection}[1]{\vspace{3pt}\noindent\textbf{#1.}}
\newcommand{\supp}{{\textbf{supplementary material}}\xspace}
\definecolor{orange}{rgb}{1,0.5,0}
\definecolor{maroon}{rgb}{0.51,0,0}

\newcommand{\res}{$\oplus$}
\newcommand{\dense}{$\rhd$\hspace{-2pt}$\lhd$}
\newcommand{\nc}{}

\newcommand{\R}[1]{{\color{black}#1}}
\newcommand{\RR}[1]{{\color{black}#1}}
\newcommand{\RRR}[1]{{\color{black}#1}}
\newcommand{\GY}[1]{{\color{gray}#1}}

\def\rot#1{\rotatebox{0}{#1}}

\begin{document}
%
\title{DeepGCNs:\\ Making GCNs Go as Deep as CNNs\small\\\url{https://www.deepgcns.org}}
%
%
%
%

\author{Guohao Li\thanks{*Equal contribution}\textsuperscript{*}\thanks{*Corresponding author: Guohao Li - guohao.li@kaust.edu.sa}\quad Matthias M\"uller\textsuperscript{*} \quad Guocheng Qian\textsuperscript{*} \quad Itzel C. Delgadillo \quad Abdulellah  Abualshour \quad \\ Ali Thabet\quad Bernard Ghanem\\
		Visual Computing Center,~ KAUST,~ Thuwal,~ Saudi Arabia\\
		{\tt\footnotesize \{guohao.li, matthias.mueller.2, guocheng.qian, itzel.delgadilloperez, abdulellah.abualshour, ali.thabet, bernard.ghanem\}@kaust.edu.sa}}

\markboth{IEEE TPAMI, Special Issue on Graphs in Vision and Pattern Analysis.}
{Shell \MakeLowercase{\textit{et al.}}: Bare Demo of IEEEtran.cls for Computer Society Journals}
%



\IEEEtitleabstractindextext{%
\begin{abstract}
Convolutional Neural Networks (CNNs) have been very successful at solving a variety of computer vision tasks such as object classification and detection, semantic segmentation, activity understanding, to name just a few. One key enabling factor for their great performance has been the ability to train very deep networks. Despite their huge success in many tasks, CNNs do not work well with non-Euclidean data, which is prevalent in many real-world applications. Graph Convolutional Networks (GCNs) offer an alternative that allows for non-Eucledian data input to a neural network. While GCNs already achieve encouraging results, they are currently limited to architectures with a relatively small number of layers, primarily due to vanishing gradients during training. This work transfers concepts such as residual/dense connections and dilated convolutions from CNNs to GCNs in order to successfully train very deep GCNs. We show the benefit of using deep GCNs (with as many as $112$ layers) experimentally across various datasets and tasks. \RR{Specifically, we achieve very promising performance in part segmentation and semantic segmentation on point clouds and in node classification of protein functions across biological protein-protein interaction (PPI) graphs.} We believe that the insights in this work will open avenues for future research on GCNs and their application to further tasks not explored in this paper. The source code for this work is available at \href{https://github.com/lightaime/deep_gcns_torch}{https://github.com/lightaime/deep\_gcns\_torch} and \href{https://github.com/lightaime/deep_gcns}{https://github.com/lightaime/deep\_gcns} for PyTorch and TensorFlow implementation respectively.
\end{abstract}

\begin{IEEEkeywords}
Graph Convolution Network, Non-euclidean Data, 3D Semantic Segmentation, Node classification, Deep Learning
\end{IEEEkeywords}}

\maketitle

\IEEEdisplaynontitleabstractindextext

%
\IEEEpeerreviewmaketitle


\IEEEraisesectionheading{\section{Introduction}\label{sec:introduction}}

%
%
%
%

\IEEEPARstart{G}{CNs} have become a prominent research topic in recent years. There are several reasons for this trend, but above all, GCNs promise a natural extension of CNNs to non-Euclidean data. While CNNs are very powerful when dealing with grid-like structured data, \eg images,  their performance on more irregular data, \eg point clouds, graphs, \etc, is sub-par. Since many real-world applications need to leverage such data, GCNs are a very natural fit. There has already been some success in using GCNs to predict individual relations in social networks \cite{social_tang2009relational}, model proteins for drug discovery \cite{chem_zitnik2017predicting,chem_wale2008comparison}, enhance predictions of recommendation engines \cite{rec_monti2017geometric,rec_ying2018graph}, and efficiently segment large point clouds \cite{wang2018dynamic}. While these works show promising results, they rely on simple and shallow network architectures.

In the case of CNNs, the primary reason for their continued success and state-of-the-art performance on many computer vision tasks, is the ability to reliably train very deep network architectures. Surprisingly, it is not clear how to train deep GCN architectures and many existing works have investigated this limitation along with other shortcomings of GCNs \cite{li2018deeper,wu2019comprehensive,zhou2018graph}. Similar to CNNs, stacking multiple layers in GCNs leads to the vanishing gradient problem. 
The over-smoothing problem can also occurs when repeatedly applying many GCN layers \cite{li2018deeper}. In this case, it was observed that the features of vertices within each connected component will converge to the same value and thus become indistinguishable from each other. As a result, most state-of-the-art GCNs are limited to shallow network architectures, usually no deeper than $4$ layers \cite{zhou2018graph}.

\begin{figure}[!th]
    \centering
    \begin{tabular}{cc}
    \includegraphics[trim=5mm 0mm 0mm 0mm, width=1\columnwidth]{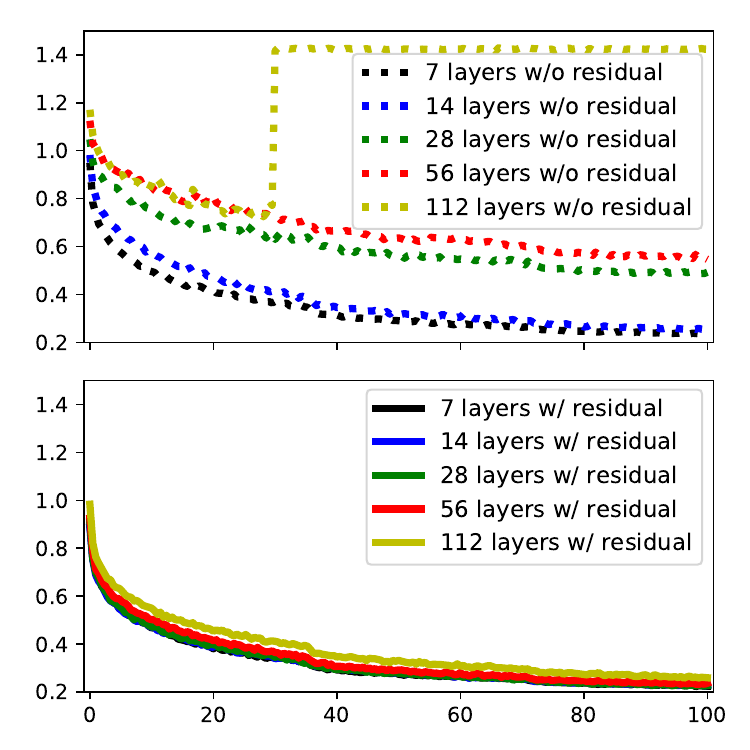}
    \end{tabular}
    \caption{\textbf{Training DeepGCNs}. (\RR{\emph{top}}) We show the square root of the training loss for GCNs with 7, 14, 28, 56 and 112 layers, with and without residual connections for $100$ epochs. We note that adding more layers without residual connections translates to a substantially higher loss and for very deep networks (\eg 112 layers) even to divergence. (\RR{\emph{bottom}}) In contrast, training GCNs with residual connections results in consistent training stability across all depths. \R{All training losses are obtained by training ResGCNs with varying depth but the same hyperparameters for the task of semantic segmentation on the S3DIS dataset.}}
    \label{fig:intro_fig}
\end{figure}

The vanishing gradient problem is well-known and well-studied in the realm of CNNs. As a matter of fact, it was the key limitation for deep convolutional networks before ResNet \cite{he2016deep} proposed a simple, yet effective solution. The introduction of residual connections \cite{he2016deep} between consecutive layers addressed the vanishing gradient problem by providing additional paths for the gradient.
This enabled deep residual networks with more than a hundred layers (\eg ResNet-152) to be trained reliably. The idea was further extended by DenseNet \cite{huang2017densely}, where additional connections are added across layers. 

Training deep networks reveals another bottleneck, which is especially relevant for tasks that rely on the spatial composition of the input image, \eg object detection, semantic segmentation, depth estimation, \etc With increased network depth, more spatial information can potentially be lost during pooling. Ideally, the receptive field should increase with network depth without loss of resolution. For CNNs, dilated convolutions \cite{yu2015multi} were introduced to tackle this issue. The idea is again simple but effective. Essentially, the convolutions are performed across more distant neighbors as the network depth increases. In this way, multiple resolutions can be seamlessly encoded in deeper CNNs. Several innovations, in particular residual/dense connections and dilated convolutions, have enabled reliable training of very deep CNNs achieving state-of-the-art performance on many tasks. This yields the following question: do these innovations have a counterpart in the realm of GCNs?

In this work, we present an extensive study of methodologies that allow training very deep GCNs. We adapt concepts that were successful in training deep CNNs, in particular residual connections, dense connections, and dilated convolutions. We show how these concepts can be incorporated into a graph framework. In order to quantify the effect of these additions, we conduct an extensive analysis of each component and its impact on accuracy and stability of deep GCNs. To showcase the potential of these concepts in the context of GCNs, we apply them to the popular tasks of semantic segmentation and part segmentation of point clouds as well as node classification of biological graphs. Adding either residual or dense connections in combination with dilated convolutions, enables successful training of GCNs with a depth of $112$ layers (refer to \figLabel \ref{fig:intro_fig}). \RR{The proposed deep GCNs improve the baseline model on the challenging point cloud dataset S3DIS \cite{2017arXiv170201105A} by $3.9\%$ mIOU and outperform previous methods in many classes of PartNet \cite{Mo_2019_CVPR}.} The same deep GCN architecture achieves an F1 score of $99.43$ on the very different PPI dataset \cite{chem_zitnik2017predicting}.

\vspace{6pt}\mysection{Contributions} The contributions of this work are three fold. \textbf{(1)} We adapt residual connections, dense connections, and dilated convolutions, which were introduced for CNNs to enable deep GCN architectures, denoted DeepGCNs. \textbf{(2)} We present extensive experiments on point cloud and biological graph data, showing the effect of each component to the stability and performance of training deep GCNs.  We use semantic segmentation and part segmentation on point clouds, as well as, node classification of biological networks as our experimental testbeds. \textbf{(3)} We show how these new concepts enable successful training of a $112$-layer GCN, the deepest GCN architecture by a large margin. With only 28 layers, we already improve the previous best performance by almost $4\%$ in terms of mIOU on the S3DIS dataset \cite{2017arXiv170201105A}; we also outperform previous methods in the task of part segmentation for many classes of PartNet \cite{Mo_2019_CVPR}. \RR{Similarly, we achieve superior results on the PPI dataset \cite{chem_zitnik2017predicting} in the task of node classification in biological networks.}

A preliminary version of this work was published in  \cite{li2019deepgcns}. This journal manuscript extends the initial version in several aspects. \emph{First}, we investigate even deeper \emph{DeepGCN} architectures with more than 100 layers. Interestingly, we find that divergence occurs when training a \emph{PlainGCN} with 112 layers, while our proposed counterpart, \emph{ResGCN} with skip connections and dilated convolutions, converges without problem (see \figLabel \ref{fig:intro_fig}).
\emph{Second}, to investigate the generality of our \emph{DeepGCN} framework, we perform extensive additional experiments on the tasks of part segmentation on PartNet and node classification on PPI. \emph{Third}, we examine the performance and efficiency of \emph{MRGCN}, a memory-efficient GCN aggregator we propose, with thorough experiments on the PPI dataset. \RR{Our results show that \emph{DeepMRGCN} models are able to outperform previous methods.} We also demonstrate that \emph{MRGCN} is very memory-efficient compared to other GCN operators via GPU memory usage experiments. \emph{Finally}, to ensure the reproducibility of our experiments and contribute to the graph learning research community, we have published code for training, testing, and visualization along with several pretrained models in both TensorFlow and PyTorch. \RR{To the best of our knowledge, our work is the first to successfully train deep GCNs beyond 100 layers and achieves superb results on both point cloud and biological graph data.}





\section{Related Work}
\label{sec:related}
A large number of real-world applications deal with non-Euclidean data, which cannot be systematically and reliably processed by CNNs in general. To overcome the shortcomings of CNNs, GCNs provide well-suited solutions for non-Euclidean data processing, leading to greatly increasing interest in using GCNs for a variety of applications. In social networks \cite{social_tang2009relational}, graphs represent connections between individuals based on mutual interests/relations. These connections are non-Euclidean and highly irregular. GCNs help better estimate edge strengths between the vertices of social network graphs, thus leading to more accurate connections between individuals. Graphs are also used to model chemical molecule structures \cite{chem_zitnik2017predicting,chem_wale2008comparison}. Understanding the bio-activities of these molecules can have substantial impact on drug discovery. Another popular use of graphs is in recommendation engines \cite{rec_monti2017geometric,rec_ying2018graph}, where accurate modelling of user interactions leads to improved product recommendations. Graphs are also popular modes of representation in natural language processing \cite{nlp_bastings2017graph,nlp_marcheggiani2017encoding}, where they are used to represent complex relations between large text units. 

GCNs also find many applications in computer vision. In scene graph generation, semantic relations between objects are modelled using a graph. This graph is used to detect and segment objects in images, and also to predict semantic relations between object pairs \cite{qi20173d,cv_scene_xu2017scene,cv_scene_yang2018graph,cv_scene_li2018factorizable}. Scene graphs facilitate the inverse process as well, where an image is reconstructed given a graph representation of the scene \cite{cv_inv_scene_johnson2018image}. Graphs are also used to model human joints for action recognition in video \cite{cv_action_yan2018spatial,cv_action_jain2016structural}.  
Moreover, GCNs are a perfect candidate for 3D point cloud processing, especially since the unstructured nature of point clouds poses a representational challenge for systematic research. Several attempts in creating structure from 3D data exist by either representing it with multiple 2D views \cite{mv_su2015multi,mv_guerry2017snapnet,mv_boulch2017unstructured,mv_li2016lstm}, or by voxelization \cite{voxel_dai2017scannet,voxel_mv_qi2016volumetric,voxel_riegler2017octnet,voxel_tchapmi2017segcloud}. More recent work focuses on directly processing unordered point cloud representations \cite{pc_qi2017pointnet,pc_qi2017pointnet++, 3dsemseg_ICCVW17,pc_huang2018recurrent,pc_ye20183d}. The recent \emph{EdgeConv} method by Wang \etal \cite{wang2018dynamic} applies GCNs to point clouds. In particular, they propose a dynamic edge convolution algorithm for semantic segmentation of point clouds. The algorithm dynamically computes node adjacency at each graph layer using the distance between point features. This work demonstrates the potential of GCNs for point cloud related applications and beats the state-of-the-art in the task of point cloud segmentation. Unlike most other works, \emph{EdgeConv} does not rely on RNNs or complex point aggregation methods. 

Current GCN algorithms including \emph{EdgeConv} are limited to shallow depths. Recent works have attempted to train deeper GCNs. For instance, Kipf \etal trained a semi-supervised GCN model for node classification and showed how performance degrades when using more than 3 layers \cite{kipf2016semi}. Pham \etal \cite{pham2017column} proposed Column Network (CLN) for collective classification in relational learning, where they showed peak performance at 10 layers and degrading performance for deeper graphs. Rahimi \etal \cite{rahimi2018semi} developed a Highway GCN for user geo-location in social media graphs, where they add ``highway" gates between layers to facilitate gradient flow. Even with these gates, the authors demonstrate performance degradation after 6 layers of depth. Xu \etal \cite{xu2018representation} developed a \emph{Jump Knowledge Network} for representation learning and devised an alternative strategy to select graph neighbors for each node based on graph structure. As with other works, their network is limited to a small number of layers ($6$). Recently, Li \etal \cite{li2018deeper} studied the depth limitations of GCNs and showed that deep GCNs can cause over-smoothing, which results in features at vertices within each connected component converging to the same value. Other works \cite{wu2019comprehensive,zhou2018graph} also show the limitations of stacking GCN layers, specifically highly complex back-propagation and the common vanishing gradient problem. 

Many difficulties facing GCNs nowadays (\eg vanishing gradients and limited receptive field) were also present in the early days of CNNs \cite{he2016deep,yu2015multi}. We bridge this gap and show that the majority of these drawbacks can be remedied by borrowing several orthogonal tricks from CNNs. Deep CNNs achieved a huge boost in performance with the introduction of ResNet \cite{he2016deep}. By adding residual connections between inputs and outputs of layers, ResNet tends to alleviate the vanishing gradient problem. DenseNet \cite{huang2017densely} takes this idea a step further and adds connections across layers as well. Dilated Convolutions \cite{yu2015multi} are another recent approach that has lead to significant performance gains, specifically in image-to-image translation tasks such as semantic segmentation \cite{yu2015multi}, by increasing the receptive field without loss of resolution. In this work, we show how one can benefit from concepts introduced for CNNs, mainly residual/dense connections and dilated convolutions, to train very deep GCNs. We support our claim by extending 
different GCN variants to deeper versions through adapting these concepts, and therefore significantly increasing their performance. Extensive experiments on the tasks of semantic segmentation and part segmentation of point clouds and node classification in biological graphs validate these ideas for general graph scenarios.

\section{Methodology} 
\label{sec:methodology}

\begin{figure*}
    \centering
    \includegraphics[page=2,trim = 10mm 10mm 40mm 10mm, clip, width=\textwidth]{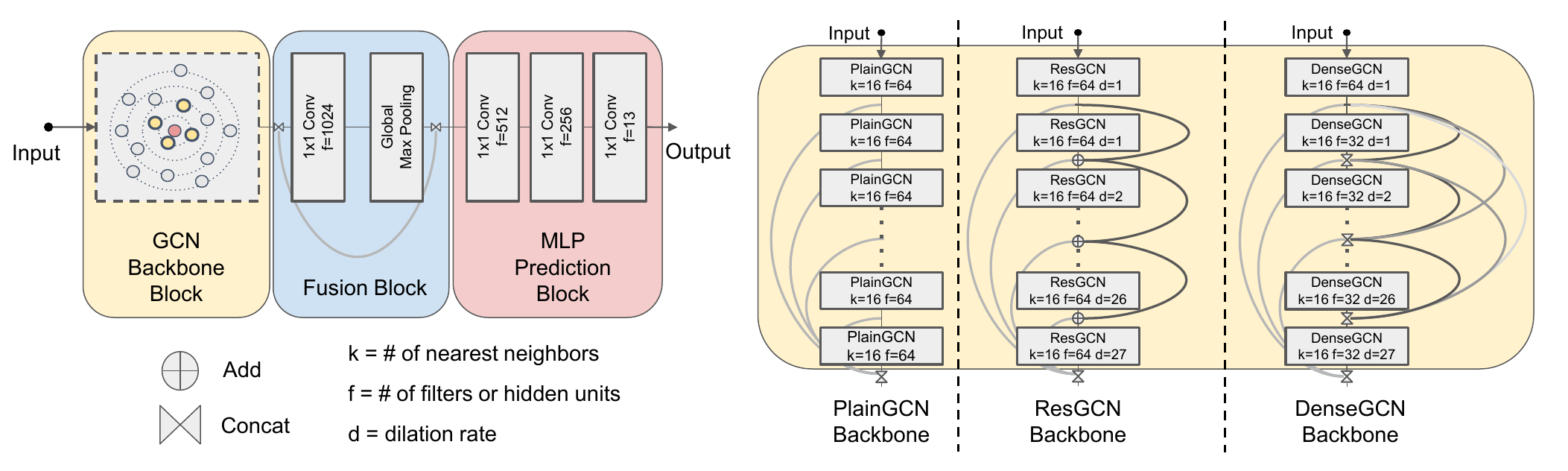}
    \caption{\textbf{Proposed GCN  architecture for point cloud semantic segmentation}. \textit{(left)} Our framework consists of three blocks: a GCN Backbone Block (feature transformation of input point cloud), a Fusion Block (global feature generation and fusion), and an MLP Prediction Block (point-wise label prediction). \textit{(right)} We study three types of GCN Backbone Block (\emph{PlainGCN}, \emph{ResGCN} and \emph{DenseGCN}) and use two kinds of layer connection (vertex-wise addition used in \emph{ResGCN} or vertex-wise concatenation used in \emph{DenseGCN}).
    }
\label{fig:pipeline}
\end{figure*}

\subsection{Representation Learning on Graphs} \label{sec:defineGCN}
\mysection{Graph Definition} A graph $\mathcal{G}$ is represented by a tuple $\mathcal{G}=(\mathcal{V}, \mathcal{E})$ where $\mathcal{V}$ is the set of unordered vertices and $\mathcal{E}$ is the set of edges representing the connectivity between vertices $v \in \mathcal{V}$. If $e_{i,j} \in \mathcal{E}$, then vertices $v_i$ and $v_j$ are connected to each other with an edge $e_{i,j}$.

\mysection{Graph Convolution Networks} Inspired by CNNs, GCNs intend to extract richer features at a vertex by aggregating features of vertices from its neighborhood. GCNs represent vertices by associating each vertex $v$ with a feature vector $\mathbf{h}_{v} \in \mathbb{R}^{D}$,
where $D$ is the feature dimension. Therefore, the graph $\mathcal{G}$ as a whole can be represented by concatenating the features of all the unordered vertices, \ie $\mathbf{h}_{\mathcal{G}} = [\mathbf{h}_{v_1}, \mathbf{h}_{v_2}, ..., \mathbf{h}_{v_N}]^{\top} \in \mathbb{R}^{N\times D}$, 
where $N$ is the cardinality of set $\mathcal{V}$. A general graph convolution operation $\mathcal{F}$ at the $l$-th layer can be formulated as the following aggregation and update operations,
\begin{equation} \label{eq:GCN}
\begin{split}
\mathcal{G}_{l+1} &=\mathcal{F}(\mathcal{G}_{l}, \mathcal{W}_l)\\
&= Update(Aggregate(\mathcal{G}_{l}, \mathcal{W}_l^{agg}), \mathcal{W}_l^{update}).
\end{split}
\end{equation}
$\mathcal{G}_l = (\mathcal{V}_{l}, \mathcal{E}_{l})$ and $\mathcal{G}_{l+1} = (\mathcal{V}_{l+1}, \mathcal{E}_{l+1})$ are the input and output graphs at the $l$-th layer, respectively.  $\mathcal{W}_l^{agg}$ and $\mathcal{W}_l^{update}$ are the learnable weights of the aggregation and update functions respectively, and they are the essential components of GCNs. In most GCN frameworks, aggregation functions are used to compile information from the neighborhood of vertices, while update functions perform a non-linear transform on the aggregated information to compute new vertex representations. There are different variants of these two functions. For example, the aggregation function can be a mean aggregator \cite{kipf2016semi}, a max-pooling aggregator \cite{pc_qi2017pointnet,hamilton2017inductive,wang2018dynamic}, an attention aggregator \cite{velivckovic2017graph}, or an LSTM aggregator \cite{peng2017cross}. The update function can be a multi-layer perceptron \cite{hamilton2017inductive,duvenaud2015convolutional}, a gated network \cite{li2015gated}, \etc More concretely, the representation of vertices is computed at each layer by aggregating features of neighbor vertices for all $v_{l+1} \in \mathcal{V}_{l+1} $ as follows,
\begin{equation} \label{GCN_node}
\resizebox{.89\columnwidth}{!} 
{$\mathbf{h}_{v_{l+1}} =\phi\left(\mathbf{h}_{v_{l}}, \rho(\{\mathbf{h}_{u_{l}}| u_{l}\in \mathcal{N}(v_{l})\}, \mathbf{h}_{v_{l}}, \mathcal{W}_{\rho}), \mathcal{W}_{\phi}\right)$},
\end{equation}
where $\rho$ is a vertex feature aggregation function and $\phi$ is a vertex feature update function, $\mathbf{h}_{v_{l}}$ and $\mathbf{h}_{v_{l+1}}$ are the vertex features at the $l$-th layer and $(l+1)$-th layer, respectively. $\mathcal{N}(v_{l})$ is the set of neighbor vertices of $v$ at the $l$-th layer, and $\mathbf{h}_{u_l}$ is the feature of those neighbor vertices parametrized by $\mathcal{W}_{\rho}$. $\mathcal{W}_{\phi}$ contains the learnable parameters of these functions. For simplicity and without loss of generality, we use a max-pooling vertex feature aggregator, without learnable parameters, to pool the difference of features between vertex $v_l$ and all of its neighbors: $\rho(.)=\max(\mathbf{h}_{u_{l}} - \mathbf{h}_{v_{l}}|~ u_{l}\in \mathcal{N}(v_{l}))$. We then model the vertex feature updater $\phi$ as a multi-layer perceptron (MLP) with batch normalization \cite{ioffe2015batch} and a ReLU as an activation function. This MLP concatenates $\mathbf{h}_{v_{l}}$ with its aggregate features from $\rho(.)$ to form its input.

\mysection{Dynamic Edges} \label{dyna} As mentioned earlier, most GCNs have fixed graph structures and only update the vertex features at each iteration. Recent work \cite{simonovsky2017dynamic,wang2018dynamic,valsesia2018learning} demonstrates that dynamic graph convolution, where the graph structure is allowed to change in each layer, can learn better graph representations compared to GCNs with fixed graph structure. For instance, ECC (Edge-Conditioned
Convolution) \cite{simonovsky2017dynamic} uses dynamic edge-conditional filters to learn an edge-specific weight matrix. Moreover, EdgeConv \cite{wang2018dynamic} finds the nearest neighbors in the current feature space to reconstruct the graph after every EdgeConv layer. In order to learn to generate point clouds, Graph-Convolution GAN (Generative Adversarial Network) \cite{valsesia2018learning} also applies $k$-NN graphs to construct the neighbourhood of each vertex in every layer. We find that dynamically changing neighbors in GCNs
results in an effectively larger receptive field, when deeper GCNs are considered. In our framework, we propose to re-compute edges between vertices via a \emph{Dilated $k$-NN} function in the feature space of each layer to further increase the receptive field. 

Designing deep GCN architectures \cite{wu2019comprehensive, zhou2018graph} is an open problem in the graph learning domain. Recent work \cite{li2018deeper, wu2019comprehensive, zhou2018graph} suggests that GCNs do not scale well to deep architectures, since stacking multiple layers of graph convolutions leads to high complexity in back-propagation. As such, most state-of-the-art GCN models are usually quite shallow \cite{zhou2018graph}. Inspired by the huge success of ResNet \cite{he2016deep}, DenseNet \cite{huang2017densely}, and Dilated Convolutions \cite{yu2015multi}, we transfer these ideas to GCNs to unleash their full potential. This enables much deeper GCNs that reliably converge in training and achieve superior performance in inference. In what follows, we provide a detailed description of three operations that can enable much deeper GCNs: residual connections, dense connections, and dilated aggregation.

\subsection{Residual Connections for GCNs} \label{sec:ResGCN}
In the original graph learning framework, the underlying mapping $\mathcal{F}$, which takes a graph as an input and outputs a new graph representation (see \eqLabel \eqref{eq:GCN}), is learned.
Here, we propose a graph residual learning framework that learns an underlying mapping $\mathcal{H}$ by fitting another mapping $\mathcal{F}$. After $\mathcal{G}_{l}$ is transformed by $\mathcal{F}$, vertex-wise addition is performed to obtain $\mathcal{G}_{l+1}$. The residual mapping $\mathcal{F}$ learns to take a graph  as input and outputs a residual graph representation $\mathcal{G}_{l+1}^{res}$ for the next layer. $\mathcal{W}_l$ is the set of learnable parameters at layer $l$. In our experiments, we refer to our residual model as \emph{ResGCN}.
\begin{equation} \label{eq2}
\begin{split}
\mathcal{G}_{l+1} &= \mathcal{H}(\mathcal{G}_{l}, \mathcal{W}_l) \\
&= \mathcal{F}(\mathcal{G}_{l}, \mathcal{W}_l) + \mathcal{G}_{l}=\mathcal{G}_{l+1}^{res}+ \mathcal{G}_{l}.
\end{split}
\end{equation}

\mysection{Dense Connections for GCNs} \label{sec:DenseGCN}
DenseNet \cite{huang2017densely} was proposed to exploit dense connectivity among layers, which improves information flow in the network and enables efficient reuse of features among layers. Inspired by DenseNet, we adapt a similar idea to GCNs so as to exploit information flow from different GCN layers. In particular, we have:
\begin{equation} \label{eq4}
\begin{split}
\mathcal{G}_{l+1} &= \mathcal{H}(\mathcal{G}_{l}, \mathcal{W}_l) \\
&= \mathcal{T}(\mathcal{F}(\mathcal{G}_{l}, \mathcal{W}_l), \mathcal{G}_{l})  \\
&= \mathcal{T}(\mathcal{F}(\mathcal{G}_{l}, \mathcal{W}_l), ...,\mathcal{F}(\mathcal{G}_{0}, \mathcal{W}_{0}), \mathcal{G}_{0}).
\end{split}
\end{equation}
The operator $\mathcal{T}$ is a vertex-wise concatenation function that densely fuses the input graph $\mathcal{G}_0$ with all the intermediate GCN layer outputs. To this end, $\mathcal{G}_{l+1}$ consists of all the GCN transitions from previous layers. Since we fuse GCN representations densely, we refer to our dense model as \emph{DenseGCN}. The growth rate of \emph{DenseGCN} is equal to the dimension $D$ of the output graph (similar to DenseNet for CNNs \cite{huang2017densely}). For example, if $\mathcal{F}$ produces a $D$ dimensional vertex feature, where the vertices of the input graph $\mathcal{G}_0$ are $D_{0}$ dimensional, the dimension of each vertex feature of $\mathcal{G}_{l+1}$ is $D_{0}+D\times (l+1)$.

\subsection{Dilated Aggregation for GCNs} \label{sec:dilation}
Dilated wavelet convolution is an algorithm originating from the wavelet processing domain \cite{holschneider1990real,shensa1992discrete}. To alleviate spatial information loss caused by pooling operations, Yu \etal \cite{yu2015multi} propose dilated convolutions as an alternative to applying consecutive pooling layers for dense prediction tasks, \eg semantic image segmentation. Their experiments demonstrate that aggregating multi-scale contextual information using dilated convolutions can significantly increase the accuracy on the semantic segmentation task. The reason behind this is the fact that dilation enlarges the receptive field without loss of resolution. We believe that dilation can also help with the receptive field of DeepGCNs. 

\begin{figure}[!htb]
    \centering
    \includegraphics[page=2,trim = 25mm 10mm 30mm 1mm, clip, width=\columnwidth]{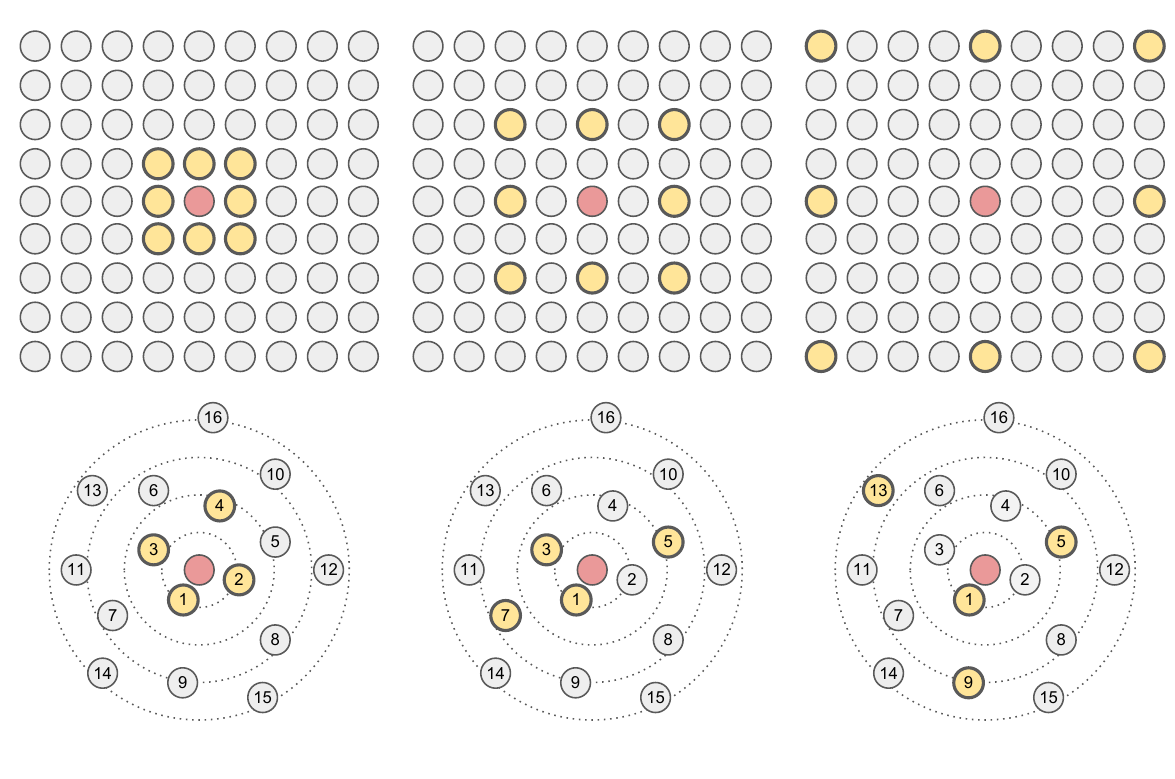}
    \caption{\textbf{Dilated Convolution in GCNs}. Visualization of dilated convolution on a structured graph arranged in a grid (\eg 2D image) and on a general structured graph. (\emph{top}) 2D convolution with kernel size 3 and dilation rate 1, 2, 4 (left to right). (\emph{bottom}) Dynamic graph convolution with dilation rate 1, 2, 4 (left to right).}
\label{fig:dilation_viz}
\end{figure}

Therefore, we introduce dilated aggregation to GCNs. There are many possible ways to construct a dilated neighborhood. We use a \emph{Dilated $k$-NN} to find dilated neighbors after every GCN layer and construct a \emph{Dilated Graph}. In particular, for an input graph $\mathcal{G}=(\mathcal{V}, \mathcal{E})$  with \emph{Dilated $k$-NN} and $d$ as the dilation rate, the \emph{Dilated $k$-NN} operation returns the $k$ nearest neighbors within the $k\times d$ neighborhood region by skipping every $d$ neighbors. The nearest neighbors are determined based on a pre-defined distance metric. In our experiments, we use the $\ell_2$ distance in the feature space of the current layer. 
Let $\mathcal{N}^{(d)}(v)$ denote the $d$-dilated neighborhood of vertex $v$. If $(u_1, u_2, ..., u_{k\times d})$ are the first sorted $k\times d$ nearest neighbors, vertices $(u_1, u_{1+d}, u_{1+2d}, ..., u_{1+(k-1)d})$ are the $d$-dilated neighbors of vertex $v$ (see \figLabel \ref{fig:dilation_viz}), \ie
\begin{equation}\label{eq5}
\mathcal{N}^{(d)}(v) = \{u_1, u_{1+d}, u_{1+2d}, ..., u_{1+(k-1)d}\}.
\end{equation}

Hence, the edges $\mathcal{E}^{(d)}$ of the output graph are defined on the set of $d$-dilated vertex neighbors $\mathcal{N}^{(d)}(v)$. Specifically, there exists a directed edge $e \in \mathcal{E}^{(d)}$ from vertex $v$ to every vertex $u \in \mathcal{N}^{(d)}(v)$. The GCN aggregation and update functions are applied, as in \eqLabel \eqref{eq:GCN}, by using the edges $\mathcal{E}^{(d)}$ created by the \emph{Dilated $k$-NN}, so as to generate the feature $\mathbf{h}_{v}^{(d)}$ of each output vertex in $\mathcal{V}^{(d)}$. We denote this layer operation as a \emph{dilated graph convolution} with dilation rate $d$, or more formally: $\mathcal{G}^{(d)}=(\mathcal{V}^{(d)}, \mathcal{E}^{(d)})$. We visualize and compare it 
to a conventional dilated convolution used in CNNs in \figLabel \ref{fig:dilation_viz}. To improve generalization, we use \emph{stochastic dilation} in practice. During training, we perform the aforementioned dilated aggregations with a high probability $(1-\epsilon)$ leaving a small probability $\epsilon$ to perform random aggregation by uniformly sampling $k$ neighbors from the set of $k\times d$ neighbors $\{u_1, u_2, ..., u_{k\times d}\}$. At inference time, we perform deterministic dilated aggregation without stochasticity, \R{\ie the dilated neighbors are sampled based on \eqLabel \ref{eq5} with probability $1$}.

\subsection{Deep GCN Variants} \label{appendix:gcn_variants}
In our experiments in the paper, we mostly work with a GCN based on EdgeConv \cite{wang2018dynamic} to show how very deep GCNs can be trained. However, it is straightforward to build other deep GCNs with the same aforementioned concepts  (\eg residual/dense graph connections and dilated graph convolutions). To show that these concepts are universal operators and can be used for general GCNs, we perform additional experiments. In particular, we build ResGCNs based on GraphSAGE \cite{hamilton2017inductive} and Graph Isomorphism Network (GIN) \cite{xu2018powerful}. In practice, we find that EdgeConv learns a better representation than the other implementations. However, it is less efficient in terms of memory and computation. Therefore, we also propose a simple GCN operation combining the advantages of both, which we refer to as \emph{MRGCN} (Max-Relative GCN). In the following, we discuss each GCN operator in detail.

\mysection{EdgeConv} Instead of aggregating neighborhood features directly, EdgeConv \cite{wang2018dynamic} proposes to first get local neighborhood information for each neighbor by subtracting the feature of the central vertex from its own feature. In order to train deeper GCNs, we add residual/dense graph connections and dilated graph convolutions to EdgeConv:
\begin{equation}
\begin{split}
\mathbf{h}^{res}_{v_{l+1}} &=\textit{max}\left(\{\textit{mlp}(\textit{concat}(\mathbf{h}_{v_{l}}, \mathbf{h}_{u_{l}}-\mathbf{h}_{v_{l}}))|u_{l}\in \mathcal{N}^{(d)}(v_{l})\}\right), \\
\mathbf{h}_{v_{l+1}} &= \mathbf{h}^{res}_{v_{l+1}} + \mathbf{h}_{v_{l}}.
\end{split}
\end{equation}

\mysection{GraphSAGE} GraphSAGE \cite{hamilton2017inductive} proposes different types of aggregator functions including a Mean aggregator, an LSTM aggregator, and a Pooling aggregator. Their experiments show that the Pooling aggregator outperforms the others. We adapt GraphSAGE with the max-pooling aggregator to obtain \emph{ResGraphSAGE}:

\begin{equation} \label{eq:sage}
\begin{split}
\mathbf{h}^{res}_{\mathcal{N}^{(d)}(v_{l})} &=\textit{max}\left(\{\textit{mlp}(\mathbf{h}_{u_{l}})|u_{l}\in \mathcal{N}^{(d)}(v_{l})\}\right), \\
\mathbf{h}^{res}_{v_{l+1}} &=\textit{mlp}\left(\textit{concat}\bigg(\mathbf{h}_{v_{l}}, \mathbf{h}^{res}_{\mathcal{N}^{(d)}(v_{l})}\bigg)\right), \\
\mathbf{h}_{v_{l+1}} &= \mathbf{h}^{res}_{v_{l+1}} + \mathbf{h}_{v_{l}},
\end{split}
\end{equation}

In the original GraphSAGE paper, the vertex features are normalized after aggregation. We implement two variants, one without normalization (see  \eqLabel \eqref{eq:sage}), another with normalization $\mathbf{h}^{res}_{v_{l+1}} = \mathbf{h}^{res}_{v_{l+1}} / \left\Vert \mathbf{h}^{res}_{v_{l+1}} \right\Vert_{2}$.

\mysection{GIN} The main difference between GIN \cite{xu2018powerful} and other GCNs is that an $\epsilon$ is learned at each GCN layer to give the central vertex and aggregated neighborhood features different weights. Hence \emph{ResGIN} is formulated as follows:
\begin{equation}
\begin{split}
\mathbf{h}^{res}_{v_{l+1}} &=\textit{mlp}\left((1+\epsilon)\cdot \mathbf{h}_{v_{l}} + \textit{sum}(\{\mathbf{h}_{u_{l}}|u_{l}\in \mathcal{N}^{(d)}(v_{l})\})\right), \\
\mathbf{h}_{v_{l+1}} &= \mathbf{h}^{res}_{v_{l+1}} + \mathbf{h}_{v_{l}}.
\end{split}
\end{equation}

\mysection{MRGCN} We find that first using a max aggregator to aggregate neighborhood relative features $(\mathbf{h}_{u_{l}} - \mathbf{h}_{v_{l}}) ,~ u_{l}\in \mathcal{N}(v_{l})$ is more effective and efficient than aggregating raw neighborhood features $\mathbf{h}_{v_{l}} ,~ u_{l}\in \mathcal{N}(v_{l})$ or aggregating features after non-linear transforms. We refer to this simple GCN as \emph{MRGCN} (Max-Relative GCN). The residual version of \emph{MRGCN} is as such:
\begin{equation}
\begin{split}
\mathbf{h}^{res}_{\mathcal{N}^{(d)}(v_{l})} &=\textit{max}\left(\{\mathbf{h}_{u_{l}}-\mathbf{h}_{v_{l}}|u_{l}\in \mathcal{N}^{(d)}(v_{l})\}\right), \\
\mathbf{h}^{res}_{v_{l+1}} &=\textit{mlp}\left(\textit{concat}\bigg(\mathbf{h}_{v_{l}}, \mathbf{h}^{res}_{\mathcal{N}^{(d)}(v_{l})}\bigg)\right), \\
\mathbf{h}_{v_{l+1}} &= \mathbf{h}^{res}_{v_{l+1}} + \mathbf{h}_{v_{l}}.
\end{split}
\end{equation}
Here, $\mathbf{h}_{v_{l+1}}$ and $\mathbf{h}_{v_{l}}$ are the hidden states of vertex $v$ at layers $l$ and $l+1$, and $\mathbf{h}^{res}_{v_{l+1}}$ is the hidden state of the residual graph. All the \emph{mlp} (multilayer perceptron) functions use a ReLU as activation function; all the \emph{max} and \emph{sum} functions above are vertex-wise feature operators; \emph{concat} functions concatenate features of two vertices into one feature vector. $\mathcal{N}^{(d)}(v_{l})$ denotes the neighborhood of vertex $v_l$ obtained from \emph{Dilated $k$-NN}. \R{\emph{MRGCN} is more efficient than \emph{EdgeConv} and \emph{GraphSAGE}. \emph{MRGCN} only applies an \emph{mlp} transform once for each vertex, while \emph{EdgeConv} needs to apply this transform for every neighbor of each vertex. In contrast to \emph{GraphSAGE}, \emph{MRGCN} does not need to apply \emph{mlp} transforms to the neighborhood features before \emph{max} aggregation and the additional computation of subtraction is negligible.}
\section{Experiments on 3D Point Clouds}\label{sec:experiments}
We propose \emph{ResGCN} and \emph{DenseGCN} to handle the vanishing gradient problem of GCNs. To enlarge the receptive field, we define a dilated graph convolution operator for GCNs. To evaluate our framework, we conduct extensive experiments on the tasks of semantic segmentation and part segmentation on large-scale 3D point cloud datasets and demonstrate that our methods significantly improve performance. In addition, we perform a comprehensive ablation study to show the effect of different components of our framework.

\subsection{Graph Learning on 3D Point Clouds}
\label{sec:PCGCN}
Point cloud segmentation is a challenging task because of the unordered and irregular structure of 3D point clouds. Normally, each point in a point cloud is represented by its 3D spatial coordinates and possibly auxiliary features such as color and/or surface normal. We treat each point as a vertex $v$ in a directed graph $\mathcal{G}$ and we use $k$-NN to construct the directed dynamic edges between points at every GCN layer (refer to \secLabel \ref{dyna}). In the first layer, we construct the input graph $\mathcal{G}_{0}$ by applying a dilated $k$-NN search to find the nearest neighbors in 3D coordinate space. At subsequent layers, we dynamically build the edges using dilated $k$-NN in feature space. For the segmentation task, we predict the categories of all the vertices at the output layer. 

\subsection{Experimental Setup}
We use the overall accuracy (OA) and  mean intersection over union (mIoU) across all classes as evaluation metrics. For each class, the IoU is computed as $\frac{TP}{TP+T-P}$, where $TP$ is the number of true positive points, $T$ is the number of ground truth points of that class, and $P$ is the number of predicted positive points. We perform the majority of our experiments on semantic segmentation of point clouds on the S3DIS dataset. To motivate the use of DeepGCNs, we do a thorough ablation study on area 5 of this dataset to analyze each component and provide insights. We then evaluate our proposed reference model \emph{ResGCN-28} (backbone of 28 layers with residual graph connections and stochastic dilated graph convolutions) on all 6 areas and compare it to the shallow DGCNN baseline \cite{wang2018dynamic} and other state-of-the-art methods. In order to validate that our method is general and does not depend on a specific dataset, we also show results on PartNet for the task of part segmentation of point clouds.

\subsection{Network Architectures}
As shown in \figLabel \ref{fig:pipeline}, all the network architectures in our experiments have three blocks: a GCN backbone block, a fusion block and an MLP prediction block. The GCN backbone block is the only part that differs between experiments. For example, the only difference between \emph{PlainGCN} and \emph{ResGCN} is the use of residual skip connections for all GCN layers in \emph{ResGCN}. Both have the same number of parameters. We linearly increase the dilation rate $d$ of dilated $k$-NN with network depth. For fair comparison, we keep the fusion and MLP prediction blocks the same for all architectures. 
The GCN backbone block takes as input a point cloud with 4096 points, extracts features by applying consecutive GCN layers to aggregate local information, and outputs a learned graph representation with 4096 vertices. The fusion and MLP prediction blocks follow a similar architecture as PointNet \cite{pc_qi2017pointnet} and DGCNN \cite{wang2018dynamic}. The fusion block is used to fuse the global and multi-scale local features. It takes as input the extracted vertex features from the GCN backbone block at every GCN layer and concatenates those features, then passes them through a 1$\times$1 convolution layer followed by max pooling. The latter layer aggregates the vertex features of the whole graph into a single global feature vector, which in return is concatenated with the feature of each vertex from all previous GCN layers (fusion of global and local information). The MLP prediction block applies three MLP layers to the fused features of each vertex/point to predict its category. In practice, these layers are  1$\times$1 convolutions.

\mysection{PlainGCN} This baseline model consists of a \emph{PlainGCN} backbone block, a fusion block, and an MLP prediction block.
The backbone stacks EdgeConv \cite{wang2018dynamic} layers with dynamic $k$-NN, each of which is similar to the one used in DGCNN \cite{wang2018dynamic}. 
No skip connections are used here. 

\mysection{ResGCN} We construct \emph{ResGCN} by adding dynamic dilated $k$-NN and residual graph connections to \emph{PlainGCN}. The connections between all GCN layers in the GCN backbone block do not increase the number of parameters.

\mysection{DenseGCN} Similarly, \emph{DenseGCN} is built by adding dynamic dilated $k$-NN and dense graph connections to the \emph{PlainGCN}. As described in \secLabel \ref{sec:DenseGCN}, dense graph connections are created by concatenating all the intermediate graph representations from previous layers. The dilation rate schedule of our \emph{DenseGCN} is the same as for \emph{ResGCN}.

\subsection{Implementation}
For semantic segmentation on S3DIS \cite{2017arXiv170201105A}, we implement our models using TensorFlow, and for part segmentation on PartNet \cite{Mo_2019_CVPR}, we implement them using PyTorch. For fair comparison, we use the Adam optimizer with the same initial learning rate $0.001$ and the same learning rate schedule for all experiments; the learning rate decays $50\%$ every $3\times 10^5$ gradient decent steps. Batch normalization is applied to every layer. Dropout with a rate of $0.3$ is used at the second MLP layer of the MLP prediction block. As mentioned in \secLabel \ref{sec:dilation}, we use dilated $k$-NN with a random uniform sampling probability $\epsilon=0.2$ for GCNs with dilations. In order to isolate the effect of the proposed DeepGCN architectures, we do not use any data augmentation or post processing techniques. We train our models end-to-end from scratch for $100$ epochs. We evaluate every $10^{\text{th}}$ epoch on the test set and report the best result for each model. The networks are trained with two NVIDIA Tesla V100 GPUs using data parallelism in semantic segmentation on S3DIS, and the batch size is set to $8$ for each GPU. For part segmentation on PartNet, we set the batch size to $7$ and the networks are trained with one NVIDIA Tesla V100 GPU.

\begin{table*}[!htb]
\centering
\footnotesize 
\setlength{\tabcolsep}{5pt} 
\begin{tabular}{c|lcc|cccccccc}
\toprule
\textbf{\rot{Ablation}} & \textbf{\rot{Model}} & \textbf{\rot{Operator}} & \textbf{\rot{mIoU}} & \textbf{\rot{$\Delta$mIoU}} & \textbf{\rot{dynamic}} & \textbf{\rot{connection}} & \textbf{\rot{dilation}} & \R{\textbf{\rot{sto. eps.}}} & \textbf{\rot{\# NNs}} & \textbf{\rot{\# filters}} & \textbf{\rot{\# layers}} \\
\midrule
\textbf{Reference}                 & \emph{ResGCN-28} & EdgeConv & \textbf{52.49} & 0.00 & \checkmark  & \res & \checkmark & 0.2 & 16 & 64 & 28 \\
\midrule
\multirow{7}{*}{\textbf{Dilation}} &        & EdgeConv & 51.98 & -0.51 & \checkmark  & \res & \checkmark &  0.0  & 16 & 64 & 28 \\
                                   &        & \R{EdgeConv} & \R{52.98} & \R{0.49} & \checkmark  & \res & \checkmark &  \R{0.4}  & 16 & 64 & 28 \\
                                   &        & \R{EdgeConv} & \R{51.92} & \R{-0.57} & \checkmark  & \res & \checkmark &  \R{0.6}  & 16 & 64 & 28 \\
                                   &        & \R{EdgeConv} & \R{52.35} & \R{-0.14} & \checkmark  & \res & \checkmark &  \R{0.8}  & 16 & 64 & 28 \\
                                   &        & \R{EdgeConv} & \R{51.40} & \R{-1.08} & \checkmark  & \res & \checkmark &  \R{1.0}  & 16 & 64 & 28 \\
                                   &        & EdgeConv & 49.64 & -2.85 & \checkmark  & \res &            &            & 16 & 64 & 28 \\ 
                                   & \emph{PlainGCN-28} & EdgeConv & \textbf{40.31} & -12.18 & \checkmark  & \nc &            &            & 16 & 64 & 28 \\
\midrule
\multirow{2}{*}{\RR{\textbf{Downsampling}}}  & \RR{\emph{ResGCN-28-FPS}}& EdgeConv & 47.98 & -4.51 & \checkmark & \res &            & 0.2 & 16 & 64 & 28 \\   
& \RR{\emph{ResGCN-28-random}}& EdgeConv & 44.30 & -8.19 & \checkmark & \res &            & 0.2 & 16 & 64 & 28 \\  
\midrule
\multirow{2}{*}{\textbf{Fixed $k$-NN}}  & & EdgeConv & 48.38 & -4.11 &             & \res &            &            & 16 & 64 & 28 \\   
                                          & & EdgeConv & 43.43 & -9.06 &             & \nc     &            &            & 16 & 64 & 28 \\  
\midrule
\multirow{6}{*}{\textbf{Connections}} & \emph{DenseGCN-28} & EdgeConv & \textbf{51.27} & -1.22 & \checkmark  & \dense    & \checkmark & 0.2 &  8 & 32 & 28 \\
                                      & \emph{\R{ResGCN-$28^\dagger$}} & \R{EdgeConv} & \R{50.86} & \R{-1.63} & \checkmark  & \res    & \checkmark & 0.2 &  16 & 64 & 28 \\
                                      &     & EdgeConv & 40.47 & -12.02 & \checkmark  & \nc     & \checkmark & 0.2 & 16 & 64 & 28 \\
                                      &     & EdgeConv & 38.79 & -13.70 & \checkmark  & \nc     & \checkmark & 0.2 &  8 & 64 & 56 \\
                                      &     & EdgeConv & 49.23 & -3.26 & \checkmark  & \nc     & \checkmark & 0.2 & 16 & 64 & 14 \\
                                      &     & EdgeConv & 47.92 & -4.57 & \checkmark  & \nc     & \checkmark & 0.2 & 16 & 64 & 7 \\ 
\midrule
\multirow{3}{*}{\textbf{Neighbors}}&        & EdgeConv & 49.98 & -2.51  & \checkmark & \res & \checkmark & 0.2 & 8 & 64 & 28 \\
                                   &        & EdgeConv & 49.22 & -3.27  & \checkmark & \res & \checkmark & 0.2 & 4 & 64 & 28 \\
                                &           & EdgeConv & 49.18 & -3.31 & \checkmark  & \res & \checkmark & 0.2 & 32 & 32 & 28 \\
\midrule
\multirow{4}{*}{\textbf{Depth}} 
                                & \emph{ResGCN-112}          & EdgeConv & 51.97 & -0.52 & \checkmark  & \res & \checkmark & 0.2 & 4 & 64 & 112 \\
                                & \emph{ResGCN-56}          & EdgeConv & \textbf{53.64} & 1.15 & \checkmark  & \res & \checkmark & 0.2 & 8 & 64 & 56 \\
                                & \emph{ResGCN-14}          & EdgeConv & 49.90 & -2.59 & \checkmark  & \res & \checkmark & 0.2 & 16 & 64 & 14 \\
                                & \emph{ResGCN-7}          & EdgeConv & 48.95 & -3.53 & \checkmark  & \res & \checkmark & 0.2 & 16 & 64 & 7 \\
\midrule
\multirow{3}{*}{\textbf{Width}} & \emph{ResGCN-28W} & EdgeConv & \textbf{53.78} & 1.29 & \checkmark  & \res & \checkmark & 0.2 & 8 & 128 & 28 \\
                                &           & EdgeConv & 48.80 & -3.69 & \checkmark  & \res & \checkmark & 0.2 & 16 & 32 & 28 \\
                                &           & EdgeConv & 45.62 & -6.87 & \checkmark  & \res & \checkmark & 0.2 & 16 & 16 & 28 \\
\midrule
\multirow{4}{*}{\textbf{GCN Variants}} &  & GraphSAGE & {49.20} & -3.29 & \checkmark  & \res & \checkmark & 0.2 & 16 & 64 & 28 \\
                                       &  & GraphSAGE-N & {49.02} & -3.47 & \checkmark  & \res & \checkmark & 0.2 & 16 & 64 & 28 \\
                                       &  & GIN-$\epsilon$ & {42.81} & -9.68 & \checkmark  & \res & \checkmark & 0.2 & 16 & 64 & 28 \\
                                       & \emph{ResMRGCN-28} & MRGCN & \textbf{51.17} & -1.32 & \checkmark  & \res & \checkmark & 0.2 & 16 & 64 & 28 \\
\bottomrule
\end{tabular}
\vspace{3pt}
\caption{
\RR{\textbf{Ablation study on area 5 of S3DIS}.} We compare our reference network (\emph{ResGCN-28}) with 28 layers, residual graph connections, and dilated graph convolutions to several ablated variants. All models were trained with the same hyper-parameters for 100 epochs on all areas except for area 5, which is used for evaluation. We denote residual and dense connections with the \res~ and \dense~ symbols respectively. \R{$\dagger$ denotes that residual connections are added between every two GCN layers.} We highlight the most important results in bold. $\Delta$mIoU denotes the difference in mIoU with respect to the reference model \emph{ResGCN-28}.}
\label{tbl:ablation_area5}
\end{table*}

\subsection{Results}
For convenient referencing, we use the naming convention \emph{BackboneBlock-\#Layers} to denote the key models in our analysis. We focus on residual graph connections for our analysis, since \emph{ResGCN-28} is easier and faster to train, but we expect that our observations also hold for dense graph connections.

\subsubsection{Semantic Segmentation on S3DIS}
In order to thoroughly evaluate the ideas proposed in this paper, we conduct extensive experiments on the Stanford large-scale 3D Indoor Spaces Dataset (S3DIS), a large-scale indoor dataset for 3D semantic segmentation of point clouds. S3DIS covers an area of more than $6,000 m^2$ with semantically annotated 3D meshes and point clouds. In particular, the dataset contains 6 large-scale indoor areas represented as colored 3D point clouds with a total of 695,878,620 points. As is common practice, we train networks on 5 out of the 6 areas and evaluate them on the left out area. 

\begin{figure} [!htb]
    \centering
    \includegraphics[trim = 2mm 0mm 0mm 0mm, clip, width=\columnwidth]{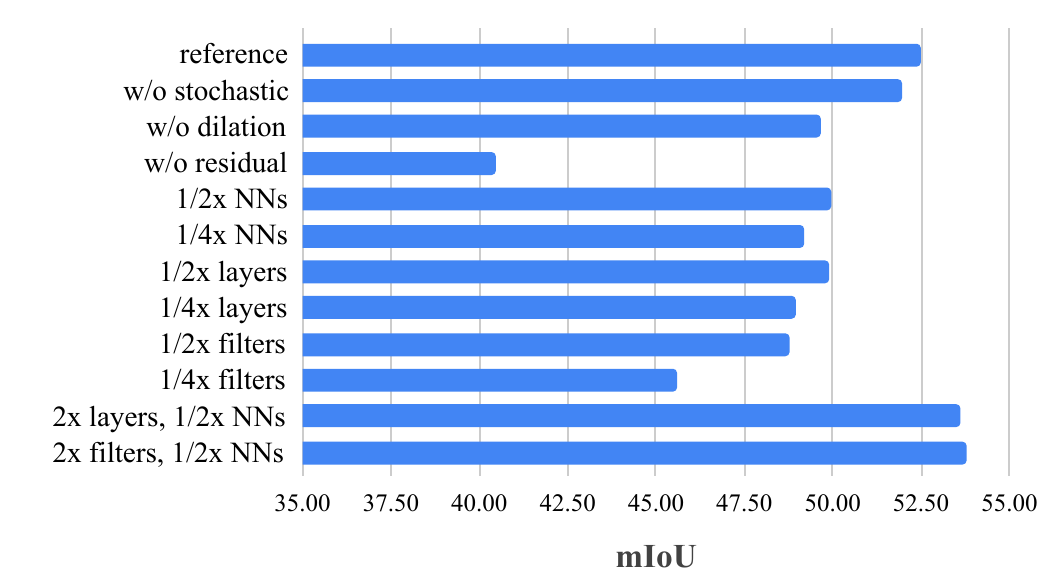}
    \caption{\textbf{Ablation study on area 5 of S3DIS}. We compare our reference network (\emph{ResGCN-28}) with 28 layers, residual graph connections, and dilated graph convolutions with several ablated variants. All models were trained  with the same hyper-parameters for 100 epochs on all areas except for area 5 of the S3DIS dataset.}
    \label{fig:ablation}
\end{figure}

We begin with the extensive ablation study, where we evaluate the trained models on area 5, after training them on the other areas. Our aim is to shed light on the contribution of each component of our novel network architectures. To this end, we investigate the performance of different \textit{ResGCN} architectures, \eg with dynamic dilated $k$-NN, with regular dynamic $k$-NN (without dilation), and with fixed edges. We also study the effect of different parameters, \eg number of $k$-NN neighbors (4, 8, 16, 32), number of filters (32, 64, 128), and number of layers (7, 14, 28, 56). To ensure that our contributions (residual/dense connections and dilated graph convolutions) are general, we apply them to multiple GCN variants that have been proposed in the literature. Overall, we conduct 25 experiments and report detailed results in \tblLabel \ref{tbl:ablation_area5}. We also summarize the most important insights of the ablation study in \figLabel \ref{fig:ablation}. In the following, we discuss each block of experiments. 

\begin{figure} [!htb]
    \centering
    \includegraphics[trim = 5mm 0mm 5mm 0mm, clip, width=\columnwidth]{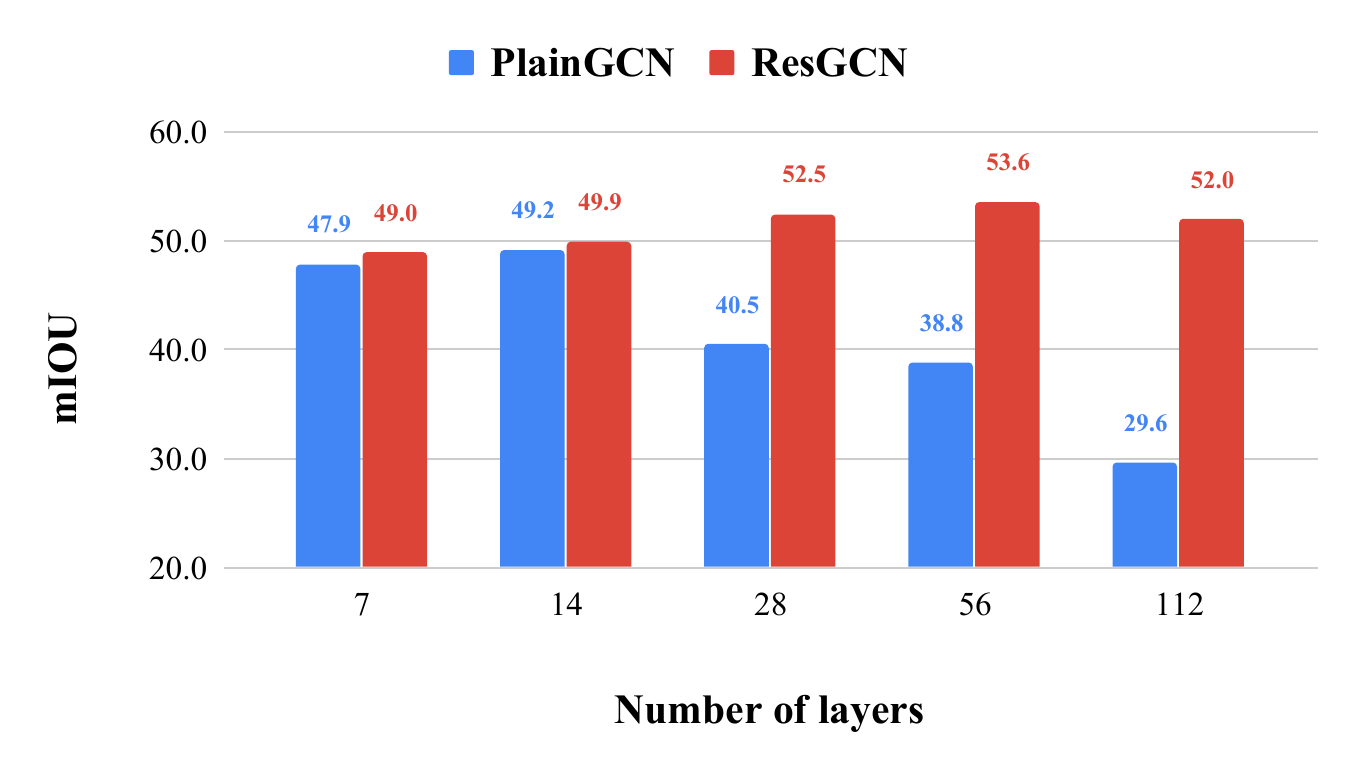}\vspace{-12pt}
    \caption{\textbf{PlainGCN vs. ResGCN on area 5 of S3DIS}. We compare networks of different depths with and without \emph{residual graph connections}. All models were trained for 100 epochs on all areas except for area 5 with the same hyper-parameters. Only when residual graph connections are used do the results improve with increasing depth.}
    \label{fig:plainVSres}
\end{figure}

\mysection{Effect of residual graph connections}
Our experiments in \tblLabel \ref{tbl:ablation_area5} (\textit{Reference})
show that residual graph connections play an essential role in training deeper networks, as they tend to result in more stable gradients. This is analogous to the insight from CNNs \cite{he2016deep}. \R{ResNet \cite{he2016deep} adds skip connections between every two convolutional blocks; we conduct experiments with two skip connection variants, \emph{ResGCN-28} and \emph{ResGCN-$28^\dagger$}. \emph{ResGCN-28} adds skip connections between every GCN layer while \emph{ResGCN-$28^\dagger$} adds skip connections between every two GCN layers. We find that \emph{ResGCN-28} outperforms \emph{ResGCN-$28^\dagger$} by $1.63\%$ mIOU (refer to \tblLabel \ref{tbl:ablation_area5}). Thus, we adopt skip connections between every layer for all other ResGCN architectures.} When the residual graph connections between layers are removed (\ie in \emph{PlainGCN-28}), performance dramatically degrades (-12\% mIoU). \figLabel \ref{fig:plainVSres} shows the importance of residual graph connections very clearly. As network depth increases, skip connections become critical for convergence. We also show similar performance gains by combining residual graph connections and dilated graph convolutions with other types of GCN layers. These results can be seen in the ablation study \tblLabel \ref{tbl:ablation_area5} (\textit{GCN Variants}) and are further discussed later in this section.

\mysection{Effect of dilation}
Results in \tblLabel \ref{tbl:ablation_area5} (\textit{Dilation}) \cite{yu2015multi} show that dilated graph convolutions account for a 2.85\% improvement in mean IoU (\textit{row 3}), 
justified primarily by the expansion of the network's receptive field.
We find that adding stochasticity to the dilated $k$-NN does helps performance but not to a significant extent. \R{We also investigate the effect of $\epsilon$ for \emph{stochastic dilation} and report the results in \tblLabel \ref{tbl:ablation_area5} (\textit{sto. eps.}). We find that models are robust to different values of $\epsilon$. The mIOUs vary in the range of about $1\%$ for different $\epsilon$ values. It is possible to achieve better performance by carefully tuning $\epsilon$. For instance, \emph{ResGCN-28} with $\epsilon$ set to $0.4$ increases the mIOU by 0.49\% compared the \textit{Reference} model with $\epsilon$ equal to $0.2$.}
Interestingly, our results in \tblLabel \ref{tbl:ablation_area5}
also indicate that dilation especially helps deep networks when combined with residual graph connections (\textit{rows 1,13}). Without such connections, performance can actually degrade with dilated graph convolutions. The reason for this is probably that these varying neighbors result in `worse' gradients, which further hinder convergence when residual graph connections are not used.

\RR{
\mysection{Dilation \vs Downsampling}
Our proposed dilated graph convolution enlarges the receptive field without the need of downsampling. To show the advantage of dilated graph convolution compared to downsampling, we conduct experiments with U-Net-style \cite{ronneberger2015u}  \emph{ResGCN} models and report the results in \tblLabel \ref{tbl:ablation_area5} (\textit{downsampling}). For the U-Net variants, we insert downsampling modules into the backbone of our \emph{ResGCN-28}. The number of vertices is downsampled by one half after every 7 layers. The final features are fed into the sequential interpolation layers to gradually upsample the features back to their original resolution. We experiment with two popular downsampling methods in the point cloud literature, \ie random downsampling \cite{Hu2020RandLANetES} and farthest point sampling (FPS) \cite{pc_qi2017pointnet++}. We denote the architectures as  \emph{ResGCN-28-random} and \emph{ResGCN-28-FPS} respectively. We observe that \emph{ResGCN-28} with dilated graph convolutions outperforms \emph{ResGCN-28-random} by 4.51\% and \emph{ResGCN-28-FPS} by 9.06\% respectively, which clearly shows the advantage of dilation.
} \RRR{However, it is worth mentioning that U-Net like structures with down-sampling are used in recent SOTA methods such as KPConv \cite{Thomas2019KPConvFA} and RandLA-Net \cite{Hu2020RandLANetES} and achieve very promising results. The results of \emph{ResGCN-28-random} and \emph{ResGCN-28-FPS} could be sub-optimal. We conjecture that with more sophisticated architectural designs and hyper-parameter twists, U-Net like GCN architectures could reach more reasonable performance.}

\mysection{Effect of dynamic $k$-NN}
While we observe an improvement when updating the $k$ nearest neighbors after every layer, we would also like to point out that it comes at a relatively high computational cost. We show different variants without dynamic edges in \tblLabel \ref{tbl:ablation_area5} (\textit{Fixed $k$-NN}).

\mysection{Effect of dense graph connections}
We observe similar performance gains with dense graph connections (\emph{DenseGCN-28}) in \tblLabel \ref{tbl:ablation_area5} (\textit{Connections}). However, with a naive implementation, the memory cost is prohibitive. Hence, the largest model we can fit into GPU memory uses only $32$ filters and $8$ nearest neighbors, as compared to $64$ filters and $16$ neighbors in the case of its residual counterpart \emph{ResGCN-28}. Since the performance of these two deep GCN variants is similar, residual connections are more practical for most use cases and hence we focus on them in our ablation study. Yet, we do expect the same insights to transfer to the case of dense graph connections.

\mysection{Effect of nearest neighbors} Results in \tblLabel \ref{tbl:ablation_area5} (\textit{Neighbors}) show that a larger number of neighbors helps in general. As the number of neighbors is decreased by a factor of 2 and 4, the performance drops by 2.5\% and 3.3\% respectively. However, a large number of neighbors only results in a performance boost, if the network capacity is sufficiently large. This becomes apparent when we increase the number of neighbors by a factor of 2 and decrease the number of filters by a factor of 2 (refer to \textit{row 3} in \textit{Neighbors}).

\mysection{Effect of network depth}
\tblLabel \ref{tbl:ablation_area5} (\textit{Depth}) shows that increasing the number of layers improves network performance, but only if residual graph connections and dilated graph convolutions are used, as is clearly shown in \tblLabel \ref{tbl:ablation_area5} (\textit{Connections}).

\mysection{Effect of network width} Results in \tblLabel \ref{tbl:ablation_area5}  (\textit{Width}) show that increasing the number of filters leads to a similar increase in performance as increasing the number of layers. In general, a higher network capacity enables learning nuances necessary for succeeding in corner cases.

\mysection{GCN variants}
Our experiments in \tblLabel \ref{tbl:ablation_area5} (\textit{GCN Variants}) show the effect of using different GCN operators. The results clearly show that different deep GCN variants with residual graph connections and dilated graph convolutions converge better than \emph{PlainGCN}. Using our proposed \emph{MRGCN} operator achieves comparable performance to the \emph{ResGCN} reference model, which relies on \emph{EdgeConv}, while only using half the GPU memory. The \emph{GraphSAGE} operator performs slightly worse and our results also show that using normalization (\ie \emph{GraphSAGE-N}) is not essential. Interestingly, when using the \emph{GIN-$\epsilon$} operator, the network converges well during the training phase and has a high training accuracy but fails to generalize to the test set. This phenomenon is also observed in the original paper \cite{xu2018powerful}, in which the best performance is achieved when $\epsilon$ is set to $0$. 

\begin{figure}[htb]
    \centering
    \includegraphics[trim = 0mm 0mm 0mm 0mm, clip, width=\columnwidth]{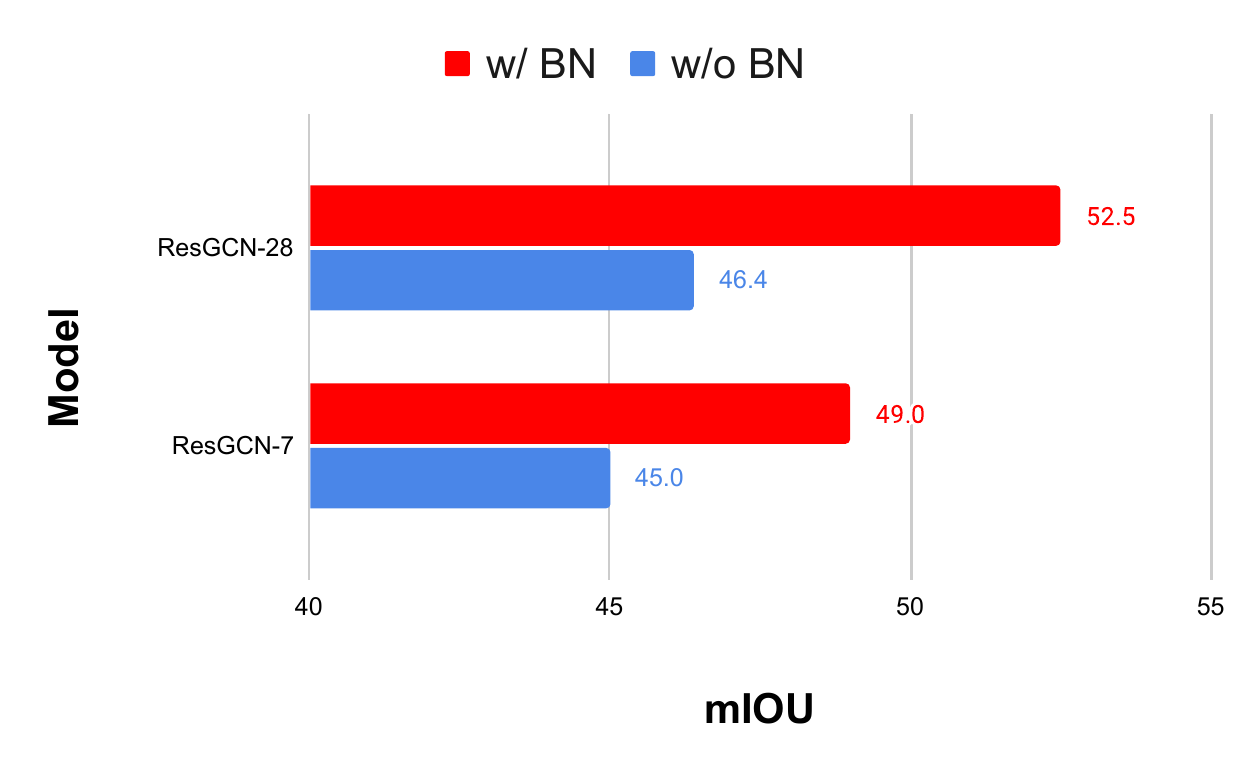}\vspace{-12pt}
    \caption{\RR{\textbf{ResGCN w/ BN \vs ResGCN w/o BN on area 5 of S3DIS}. We compare ResGCN models with and without \emph{batch normalization}. All models were trained for 100 epochs on all areas except for area 5 with the same hyper-parameters. Batch normalization is shown to be important for training deep models.}}
    \label{fig:bn_chart}
\end{figure}

\RR{
\mysection{Effect of batch normalization}
Batch normalization (BN) \cite{ioffe2015batch} is a common technique used to reduce internal covariant shift when training deep CNNs. We adopt BN in every GCN block of all the models shown in \tblLabel \ref{tbl:ablation_area5} by default. To show the effect of BN, we conduct further experiments on area 5 where we remove BN from \emph{ResGCN-28} and \emph{ResGCN-7}. The results in \figLabel \ref{fig:bn_chart} show that the performance of \emph{ResGCN-28} and \emph{ResGCN-7} drops by 6.1\% and 4\% respectively when BN is removed. \emph{ResGCN-28 w/o BN} is even outperformed by \emph{ResGCN-7 w/ BN}. This indicates that BN is essential for training deep GCNs.
}


\begin{table}[]
\centering
\begin{tabular}{lccr}
\toprule
\textbf{Model} & \multicolumn{1}{c}{Group Dis. Ratio} & \multicolumn{1}{c}{Instance Info. Gain} & \multicolumn{1}{c}{mIoU}   \\
\midrule
ResGCN-28      & 1.73                               & 0.46                              & 52.49                      \\
\midrule
w/o dilation   & 1.67                               & 0.43                              & 49.64                      \\
\midrule
w/o connection & 1.12                               & 0.01                              & 40.47 \\
\midrule
\end{tabular}
\vspace{3pt}
\caption{\RR{\textbf{Analysis of over-smoothing using the \textit{Group Distance Ratio} and the \textit{Instance Information Gain}}. We analyze the effect of \emph{residual graph connections} and \emph{dilated  graph convolutions} on the over-smoothing issue of deep GCNs by measuring group distance ratio and instance information gain. All models were trained for 100 epochs on all areas except for area 5 with the same hyper-parameters. Residual connections and dilation are shown to be effective to alleviate the over-smoothing issue.}}
\label{tbl:oversmoothing}
\end{table}

\RR{
\mysection{Analysis of over-smoothing}
To study the effect of \emph{residual graph connections} and \emph{dilated graph convolutions} on the over-smoothing issue, we adopt two quantitative metrics proposed in \cite{zhou2020towards}, \ie \emph{Group Distance Ratio} and \emph{Instance Information Gain}, to measure the over-smoothness of the learned node representation of the last GCN layer. The group distance ratio measures the ratio of the inter-group distance and the intra-group distance in the Euclidean space of the final representation. The instance information gain measures the mutual information between a input node feature and the final representation. A higher group distance ratio or instance information gain implies less over-smoothing. We measure the group distance ratio and instance information gain of the trained reference model (\emph{ResGCN-28}) and the corresponding ablated models without dilation (w/o dilation) or without residual connections (w/o connection). As shown in \tblLabel \ref{tbl:oversmoothing}, \emph{ResGCN-28} has a significantly higher group distance ratio (1.73 \vs 1.12) and instance information gain (0.46 \vs 0.01) than the non-residual counterpart. A similar observation can be made with respect to dilation albeit the impact is less pronounced. This indicates that \emph{ResGCN-28} learns sharper features and suffers less from over-smoothing.
} \RRR{However, the current over-smoothing metrics can not isolate the effects caused by the difficulty of optimization. We believe disentangling the effects of over-smoothing and vanishing gradient is a promising direction to better understand deep GCN architectures.}

\begin{figure*}[!h]
    \centering
    \includegraphics[page=1, width=\textwidth]{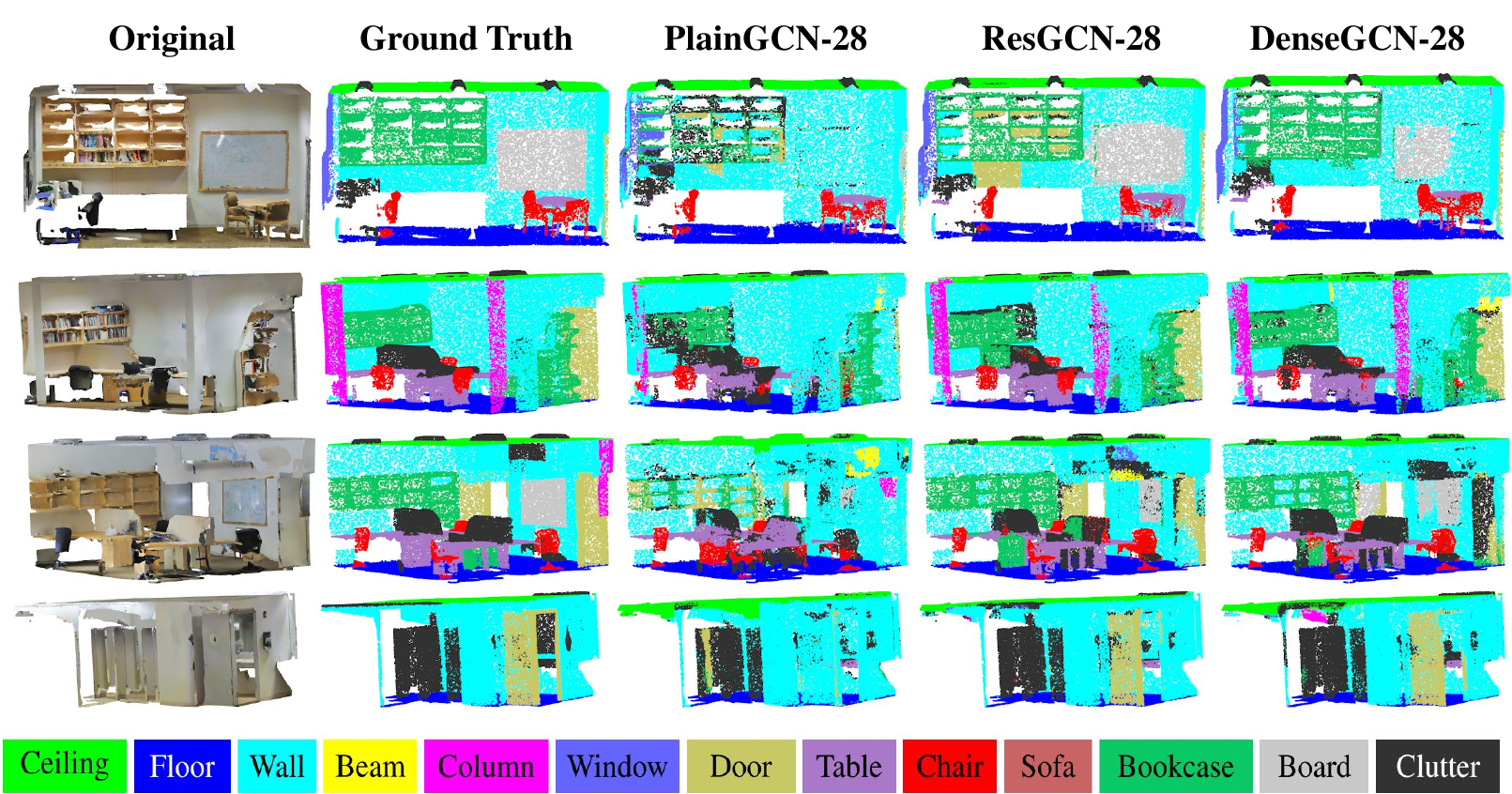}
    \caption{\textbf{Qualitative Results on S3DIS Semantic Segmentation}. We show here the effect of adding residual and dense graph connections to deep GCNs. \emph{PlainGCN-28}, \emph{ResGCN-28}, and \emph{DenseGCN-28} are identical except for the presence of residual graph connections in \emph{ResGCN-28} and dense graph connections in \emph{DenseGCN-28}. We note how both residual and dense graph connections have a substantial effect on hard classes like board, bookcase, and sofa. These are lost in the results of \emph{PlainGCN-28}.}
    \label{fig:qualitative}
\end{figure*}

\mysection{Qualitative results}
\figLabel \ref{fig:qualitative} shows qualitative results on area 5 of S3DIS \cite{2017arXiv170201105A}. As expected from the results in \tblLabel \ref{tbl:ablation_area5}, our \emph{ResGCN-28} and \emph{DenseGCN-28} perform particularly well on difficult classes such as board, beam, bookcase and door. \emph{Rows 1-4} clearly show how \emph{ResGCN-28} and \emph{DenseGCN-28} are able to segment the board, beam, bookcase and door respectively, while \emph{PlainGCN-28} completely fails. Please refer to the \supp for more qualitative results and further results.

\begin{figure*}[!h]
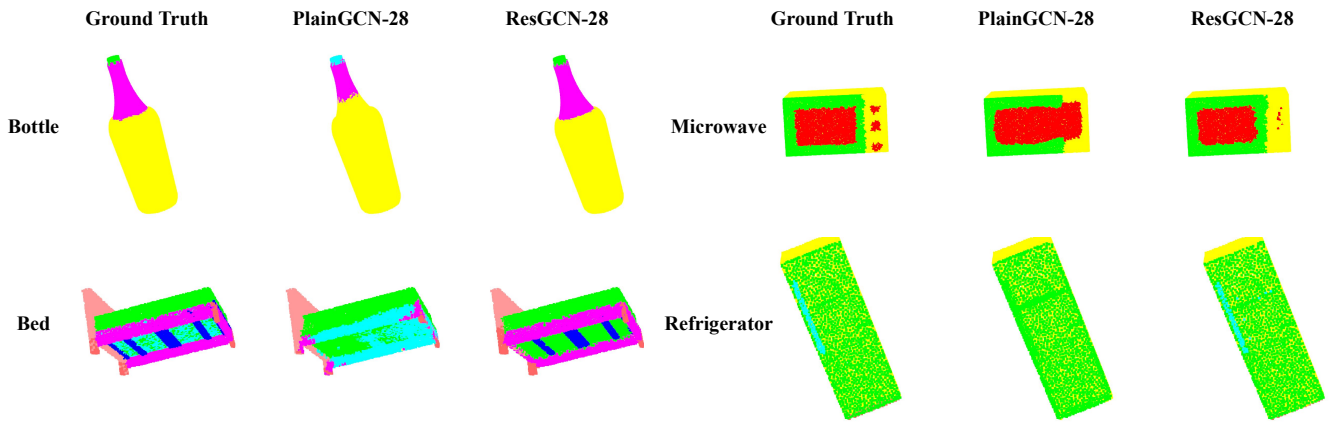

    \centering
     \includegraphics[page=1,trim = 0mm 0mm 200mm 300mm, clip, width=3.55in]{figures/PlainGCN_28_vs_ResGCN_28.pdf}
    \includegraphics[page=3,trim = 0mm 0mm 200mm 300mm, clip, width=3.55in]{figures/PlainGCN_28_vs_ResGCN_28.pdf}
    \includegraphics[page=2,trim = 0mm 0mm 200mm 600mm, clip, width=3.55in]{figures/PlainGCN_28_vs_ResGCN_28.pdf}
    \includegraphics[page=4,trim = 0mm 0mm 200mm 600mm, clip, width=3.55in]{figures/PlainGCN_28_vs_ResGCN_28.pdf}
    \caption{\textbf{Qualitative Results on PartNet Part Segmentation}. We show here the effect of adding residual connections to deep GCNs. \emph{PlainGCN-28} and \emph{ResGCN-28} are identical except for the presence of residual connections in \emph{ResGCN-28}. We note how residual connections have a positive effect on part segmentation compared to \emph{PlainGCN-28}. Many important parts of the objects are classified incorrectly using \emph{PlainGCN-28}.}
    \label{fig:qualitative_partnet}
\end{figure*}

\begin{table*}[!htb]
\centering
\footnotesize 
\setlength{\tabcolsep}{4.2pt} 
\begin{tabular}{lcc|ccccccccccccc}
\toprule
\textbf{Method}  & \textbf{OA} & \textbf{mIOU}    & \textbf{ceiling} & \textbf{floor}   & \textbf{wall}    & \textbf{beam}    & \textbf{column}  & \textbf{window}  & \textbf{door}    & \textbf{table}   & \textbf{chair}   & \textbf{sofa}    & \textbf{bookcase} & \textbf{board}   & \textbf{clutter} \\
\midrule
PointNet \cite{pc_qi2017pointnet}& 78.5    & 47.6    & 88.0    & 88.7    & 69.3    & 42.4    & 23.1    & 47.5    & 51.6    & 54.1    & 42.0    & 9.6     & 38.2    & 29.4    & 35.2    \\
MS+CU \cite{3dsemseg_ICCVW17}   & 79.2    & 47.8    & 88.6    & \textbf{95.8} & 67.3    & 36.9    & 24.9    & 48.6    & 52.3    & 51.9    & 45.1    & 10.6    & 36.8    & 24.7    & 37.5    \\
G+RCU \cite{3dsemseg_ICCVW17}    & 81.1    & 49.7    & 90.3    & 92.1    & 67.9    & \textbf{44.7} & 24.2    & 52.3    & 51.2    & 58.1    & 47.4    & 6.9     & 39.0    & 30.0    & 41.9    \\
PointNet++ \cite {pc_qi2017pointnet++} &-  &53.2 &90.2 &91.7 &73.1 &42.7 &21.2 &49.7 &42.3 &62.7 &59.0 &19.6 &45.8 &48.2 &45.6\\
3DRNN+CF \cite{pc_ye20183d}& \textbf{86.9} & 56.3    & 92.9    & 93.8    & 73.1    & 42.5    & 25.9    & 47.6    & 59.2    & 60.4    & \textbf{66.7} & 24.8    & \textbf{57.0} & 36.7    & 51.6    \\
\midrule
DGCNN \cite{wang2018dynamic}  & 84.1    & 56.1    & - & - & - & - & - & - & - & - & - & - & - & - & - \\
\textbf{ResGCN-28 (\textit{Ours})} & 85.9    & \textbf{60.0} & \textbf{93.1} & 95.3    & \textbf{78.2} & 33.9    & \textbf{37.4} & \textbf{56.1} & \textbf{68.2} & \textbf{64.9} & 61.0  & \textbf{34.6}    & 51.5    & \textbf{51.1} & \textbf{54.4}\\
\midrule
\GY{Deep LPN~\cite{Le2020GoingDW}} &\GY{85.7} &\GY{60.0} &\GY{91.0} &\GY{95.6} &\GY{76.1} &\GY{50.3} &\GY{25.9} &\GY{55.1} &\GY{56.8} &\GY{66.3} &\GY{74.3} &\GY{25.8} &\GY{54.0} &\GY{52.3} &\GY{55.3}\\
\GY{ShellNet~\cite{Zhang2019ShellNetEP}} & \GY{87.1} & \GY{66.8} & \GY{90.2} & \GY{93.6} & \GY{79.9} & \GY{60.4} & \GY{44.1} & \GY{64.9} & \GY{52.9} & \GY{71.6} & \GY{84.7} & \GY{53.8} & \GY{64.6} & \GY{48.6} & \GY{59.4} \\
\GY{RandLA-Net~\cite{Hu2020RandLANetES}} & \GY{88.0} & \GY{70.0} & \GY{93.1} & \GY{96.1} & \GY{80.6} & \GY{62.4} & \GY{48.0} & \GY{64.4} & \GY{69.4} & \GY{69.4} & \GY{76.4} & \GY{60.0} & \GY{64.2} & \GY{65.9} & \GY{60.1}\\
\GY{KPConv~\cite{Thomas2019KPConvFA}} & \GY{--} & \GY{70.6} & \GY{93.6} & \GY{92.4} & \GY{83.1} & \GY{63.9} & \GY{54.3} & \GY{66.1} & \GY{76.6} & \GY{64.0} & \GY{57.8} & \GY{74.9} & \GY{69.3} & \GY{61.3} & \GY{60.3} \\
\bottomrule
\end{tabular}
\vspace{3pt}
\caption{\textbf{Comparison of \emph{ResGCN-28} with the state-of-the-art on S3DIS Semantic Segmentation}. We report average per-class results across all areas for our reference model \emph{ResGCN-28} and state-of-the-art baselines. \emph{ResGCN-28} which has 28 GCN layers, residual graph connections, and dilated graph convolutions outperforms the previous state-of-the-art by almost $4\%$. It also outperforms all baselines in $9$ out of $13$ classes. The metrics shown are overall point accuracy (OA) and mean IoU (mIoU). '-' denotes not reported and \textbf{bold} denotes best performance.\RRR{The most recent SOTA results published after the short version of this work are reported in \GY{gray}.}}
\label{tbl:ours_vs_baselines}
\end{table*}

\mysection{Comparison to state-of-the-art} 
Finally, we compare our reference network (\emph{ResGCN-28}), which incorporates the ideas put forward in the methodology, to several state-of-the-art baselines in \tblLabel \ref{tbl:ours_vs_baselines}. The results clearly show the effectiveness of deeper models with residual graph connections and dilated graph convolutions. \emph{ResGCN-28} outperforms DGCNN \cite{wang2018dynamic} by 3.9\% (absolute) in mean IoU. DGCNN has the same fusion and MLP prediction blocks as \emph{ResGCN-28} but a shallower \emph{PlainGCN}-like backbone block.
Furthermore, we outperform all baselines in 9 out of 13 classes. We perform particularly well in the difficult object classes such as board, where we achieve 51.1\%, and sofa, where we improve state-of-the-art by about 10\% mIOU.

This significant performance improvement on the difficult classes is probably due to the increased network capacity, which allows the network to learn subtle details necessary to distinguish between a board and a wall for example. The first row in \figLabel \ref{fig:qualitative} is a representative example for this occurrence.
Our performance gains are solely due to our innovation in the network architecture, since we use the same hyper-parameters and even learning rate schedule as the baseline DGCNN \cite{wang2018dynamic} and only decrease the number of nearest neighbors from $20$ to $16$ and the batch size from $24$ to $16$ due to memory constraints. We outperform state-of-the art methods by a significant margin and expect further improvement from tweaking the hyper-parameters, especially the learning schedule. 
\RRR{In \tblLabel \ref{tbl:ours_vs_baselines}, we also include the most recent SOTA results from Deep LPN \cite{Le2020GoingDW},  ShellNet\cite{Zhang2019ShellNetEP}, RandLA-Net\cite{Hu2020RandLANetES} and KPConv\cite{Thomas2019KPConvFA} for reference. These methods employ different data pre-processing and augmentation pipelines and were published after the short version of our paper. We color these results in gray.}

\begin{table*}[!htb]
\centering
\footnotesize 
\setlength{\tabcolsep}{2.7pt} 
\begin{tabular}{l|c|cccccccccccccccccccc}
\toprule
\textbf{Method} & \R{\textbf{Avg}} & \textbf{bed} & \textbf{bottle}  & \textbf{chair} & \textbf{clock} & \textbf{dishw.} & \textbf{disp.} & \textbf{door} & \textbf{earph.} & \textbf{fauc.} & \textbf{knife} & \textbf{lamp} & \textbf{micro.} & \textbf{fridge} & \textbf{st. furn.} & \textbf{table} & \textbf{tr. can} & \textbf{vase} \\
\midrule
PointNet \cite{pc_qi2017pointnet} & 35.6& 13.4 & 29.5  & 27.8 & 28.4 & 48.9 & 76.5 & 30.4 & 33.4 & 47.6 & 32.9 & 18.9 & 37.2 & 33.5 & 38.0 & 29.0 & 34.8 & 44.4 \\ 
PointNet++ \cite{pc_qi2017pointnet++} &  42.5 & 30.3 & 41.4 & 39.2 & {41.6} & 50.1 & 80.7 & 32.6 & 38.4 & 52.4& 34.1 & {25.3} &  48.5 & 36.4 & 40.5 & {33.9} & 46.7 & 49.8 \\
SpiderCNN \cite{xu2018spidercnn}& 37.0 & 36.2 & 32.2 & 30.0 & 24.8 & 50.0 & 80.1 & 30.5 & 37.2 & 44.1& 22.2 & 19.6 & 43.9 & 39.1  & 44.6 & 20.1 & 42.4 & 32.4 \\ 
PointCNN \cite{Li2018PointCNNCO} & 46.5& {41.9}&	41.8&	{43.9}&	36.3&	58.7&	{82.5}&	{37.8}&	{48.9} &	{60.5}&	34.1&	20.1&	{58.2}&	42.9&	{49.4}&	21.3&	{53.1}&	58.9 \\
\midrule
\R{\textbf{PlainGCN-28 (\textit{Ours})}} & 31.3 & 20.1&	29.4&	24.6&	21.4&	38.2&	73.2&	24.2&	36.4&	42.2&	32.5&	16.7&	35.6&	29.8&	30.1&	15.1&	33.8&	28.2\\
\R{\textbf{ResGCN-28 (\textit{Ours})}} &45.1 & 35.9	& 49.3	& 41.1	&33.8	&56.2	&81.0	&31.1	&45.8	&52.8	&44.5	&23.1	&51.8	&34.9	&47.2	&33.6	&50.8	&54.2 \\ 
\midrule
\GY{Deep LPN ~\cite{Le2020GoingDW}} & \GY{38.6} & \GY{-} &	\GY{-}&	\GY{-}&	\GY{-}&	\GY{-}&	\GY{-}&	\GY{-}&	\GY{-} &	\GY{-}&	\GY{-}&	\GY{-}&	\GY{-}&	\GY{-}&	\GY{-}&	\GY{-}&	\GY{-}&	\GY{-} \\
\GY{PosPool \cite{Liu2020ACL}} &\GY{53.8} & \GY{-} &	\GY{-}&	\GY{-}&	\GY{-}&	\GY{-}&	\GY{-}&	\GY{-}&	\GY{-} &	\GY{-}&	\GY{-}&	\GY{-}&	\GY{-}&	\GY{-}&	\GY{-}&	\GY{-}&	\GY{-}&	\GY{-} \\
\bottomrule
\end{tabular}
\vspace{3pt}
\caption{\textbf{Comparison of \emph{ResGCN-28} with other methods on PartNet Part Segmentation.} We report the part-category mean IoU (mIoU) on the fine-grained level of segmentation. \emph{ResGCN-28} which has 28 GCN layers, residual graph connections, and dilated graph convolutions outperforms \emph{PlainGCN-28} and most previous methods w.r.t the average mIoU. \RRR{The most recent SOTA results published after the short version of this work are reported in \GY{gray}. '-' denotes that the results are not reported in the original papers.}
}  
\label{tbl:partnet_results}
\end{table*}

\subsubsection{Part Segmentation on PartNet}
We further experiment with our architecture on the task of part segmentation and evaluate it on the recently proposed large-scale PartNet \cite{Mo_2019_CVPR} dataset. PartNet consists of over 26,671 3D models from 24 object categories with 573,585 annotated part instances. The dataset establishes three benchmarking tasks for part segmentation on 3D objects: fine-grained semantic segmentation, hierarchical semantic segmentation, and instance segmentation. For the following experiments, we focus on the fine-grained level of semantic segmentation, which includes 17 out of the 24 object categories present in the PartNet dataset.

\R{We use the same reference architecture, \emph{ResGCN-28}, that we used for the experiments on the S3IDS dataset. We compare its performance to the baseline architecture \emph{PlainGCN-28} to show the impact of our residual connections and stochastic dilated convolutions. We also compare against the state-of-the-art reported in the PartNet paper \cite{Mo_2019_CVPR}, namely PointNet \cite{pc_qi2017pointnet}, PointNet++ \cite{pc_qi2017pointnet++}, SpiderCNN \cite{xu2018spidercnn}, and PointCNN \cite{Li2018PointCNNCO}. As suggested in PartNet \cite{Mo_2019_CVPR}, we use 10,000 sampled points as input. We train a separate network for each category and optimize them with Adam for $500$ epochs without weight decay. The initial learning rate is $0.005$ that is decayed by a factor of $0.9$ every $50$ epochs. We report our results using the best pretrained models on the validation sets. }

\mysection{Qualitative results} \figLabel \ref{fig:qualitative_partnet} shows qualitative results on 4 categories of PartNet \cite{Mo_2019_CVPR}: bottle, bed, microwave, and refrigerator. As expected from the results in Table \ref{tbl:partnet_results}, \emph{ResGCN-28} performs very well compared to the baseline \emph{PlainGCN-28}, where there are no residual connections between layers. Although \emph{ResGCN-28} produces some incorrect outputs compared to the ground truth in categories like microwave and bed, it still outperforms \emph{PlainGCN-28} and segments the important parts of the object. 
We provide more qualitative results in the \supp.

\mysection{Comparison to state-of-the-art}
\R{
We summarize the results of our \emph{ResGCN-28} and compare it to \emph{PlainGCN-28} and other state-of-the-art methods in \tblLabel \ref{tbl:partnet_results}. \emph{ResGCN-28} outperforms \emph{PlainGCN-28} by a large margin, which illustrates how our proposed residual connections and stochastic dilated convolutions enable training deep GCN architectures.
\emph{ResGCN-28} also substantially outperforms PointNet \cite{pc_qi2017pointnet}, PointNet++ \cite{pc_qi2017pointnet++}, and SpiderCNN \cite{xu2018spidercnn}. Note that PointCNN designs a point convolution specialized for point clouds, whereas \emph{ResGCN-28} is a general framework for any graph (\eg point clouds, biological networks, citation networks, \etc). Despite this huge disadvantage, \emph{ResGCN-28} still achieves results on par with PointCNN \cite{Li2018PointCNNCO} and even outperforms PointCNN on some categories including bottle, knife, lamp, and table. 
}
\RRR{We also include the most recent SOTA results from Deep LPN \cite{Le2020GoingDW} and PosPool \cite{Liu2020ACL} for reference. These methods employ different data pre-processing and augmentation pipelines and were published after the short version of our paper. We color these results in gray.}

\R{\subsubsection{Object Classification on ModelNet40}}
\R{We experiment with our architecture on the task of object classification and evaluate it on the popular object classification dataset ModelNet40 \cite{ben20183dmfv}. The dataset contains 12,311 CAD models from 40 different categories. Although saturated, ModelNet40 is considered an important benchmark for the task of 3D object classification. The results in \tblLabel \ref{tb: modelnet40} show the superiority of our \emph{ResGCN} and \emph{DenseGCN} architectures over the baseline \emph{PlainGCN} and state-of-the-art methods. When compared to the baseline \emph{PlainGCN}, one observes the positive effect of our residual connections, dense connections, and dilated convolutions. In addition, the \emph{ResGCN} and \emph{DenseGCN} architectures outperform all state-of-the-art approaches in overall test accuracy. Particularly, \emph{ResGCN-14} achieves 93.6\% on ModelNet-40 in terms of overall accuracy. Here, we point out that the recent work of KPConv \cite{Thomas2019KPConvFA} uses a much larger number of trainable parameters compared to our reference \emph{ResGCN-28}, as shown in \tblLabel \ref{tb: modelnet40}. And yet, even our shallower variant \emph{ResGCN-14} can outperform KPConv, while using almost one order of magnitude less trainable parameters.}

\vspace{9pt}
\begin{table}[!htb]
\centering
\setlength{\tabcolsep}{5pt} 
\begin{tabular}{l|ccc}
\toprule
\R{\textbf{Model}} & \R{\textbf{O.A. (\%)}} & \R{\textbf{C.A. (\%)}} & \R{\textbf{Param (M)}} \\ 
\midrule
\R{3DmFV-Net \cite{ben20183dmfv}} &	\R{91.6} & \R{-} & \R{45.77}\\
\R{PointNet \cite{pc_qi2017pointnet}} &	\R{89.2} & \R{86.2} & \R{3.5}\\
\R{PointNet++ \cite{pc_qi2017pointnet++}}	& \R{90.7} & \R{-} & \R{1.48}\\
\R{SO-Net \cite{Li2018SONetSN}} &	\R{90.9} & \R{87.3} & \R{-}\\
\R{PCNN by Ext \cite{Atzmon2018PointCN}}	& \R{92.3} & \R{-} & \R{8.2}\\
\R{SpecGCN \cite{Wang2018LocalSG}} & \R{91.5} & \R{-} & \R{2.05}\\
\R{SpiderCNN \cite{xu2018spidercnn}}	& \R{90.5} & \R{-} & \R{-}\\
\R{MCConv \cite{Hermosilla2018MonteCC}}	& \R{90.9} & \R{-} & \R{-}\\
\R{FlexConv \cite{Groh2018FlexConvolutionM}} &	\R{90.2} & \R{-} & \R{0.35}\\
\R{PointCNN \cite{Li2018PointCNNCO}} &	\R{92.2} & \R{88.1} & \R{0.6}\\
\R{DGCNN \cite{wang2018dynamic}} &	\R{92.9} & \R{90.2} & \R{1.84}\\
\R{KPConv rigid \cite{Thomas2019KPConvFA}} &	\R{92.9} & \R{-} & \R{14.3}\\
\R{KPConv deform \cite{Thomas2019KPConvFA}} &	\R{92.7} & \R{-} & \R{15.2}\\
\midrule
\R{\textbf{\emph{PlainGCN-14} (\textit{Ours})}} & \R{90.5} & \R{85.1}& \R{2.2}\\ 
\R{\textbf{\emph{PlainGCN-28} (\textit{Ours})}} & \R{91.4} & \R{86.8}& \R{3.3}\\ 
\R{\textbf{\emph{ResGCN-14} (\textit{Ours})}} & \textbf{\R{93.6}} & \textbf{\R{90.9}}& \R{2.2}\\ 
\R{\textbf{\emph{ResGCN-28} (\textit{Ours})}} & \R{\textbf{93.3}} &\R{89.5 }& \R{3.3}\\ 
\R{\textbf{\emph{DenseGCN-14} (\textit{Ours})}} &\R{\textbf{93.3}} &\R{89.4}& \R{8.8}\\ 
\R{\textbf{\emph{DenseGCN-28} (\textit{Ours})}} &\R{\textbf{93.2}} &\R{\textbf{90.3}}& \R{30.9}\\ 
\bottomrule
\end{tabular}
\vspace{3pt}
\caption{\R{\textbf{Comparison of our GCN variants with the state-of-the-art on ModelNet40 point cloud classification}.  We report the overall classification test accuracy (O.A.) and mean per-class accuracy (C.A.) of our models and state-of-the-art models in addition to the to total number of parameters for each model. One can observe the superiority of our \emph{ResGCN} and \emph{DenseGCN} architectures over state-of-the-art.}}
\label{tb: modelnet40}
\end{table}
\section{Experiments on Biological Networks}
\label{sec:experiments2}
In order to demonstrate the generality of our contributions and specifically our deep \emph{ResGCN} architecture, we conduct further experiments on general graph data. We choose the popular task of node classification on biological graph data, which is quite different from point cloud data.
In the following experiments, we mainly study the effects of skip connections, the number of GCN layers (\ie depth), number of filters per layer (\ie width) and different graph convolutional operators.

\subsection{Graph Learning on Biological Networks}
\label{sec:exp2-PPIGCN}
The main difference between biological networks and point cloud data is that biological networks have inherent edge information and high dimensional input features.
For the graph learning task on biological networks, we use the PPI \cite{chem_zitnik2017predicting} dataset to evaluate our architectures. PPI is a popular dataset for multi-label node classification, containing 24 graphs with 20 in the training set, 2 in the validation set, and 2 in the testing set. 
Each graph in PPI corresponds to a different human tissue, each node in a graph represents a protein and edges represent the interaction between proteins.
Each node has positional gene sets, motif gene sets, and immunological signatures as input features (50 in total) and 121 gene ontology sets as labels. The input of the task is a graph that contains 2373 nodes on average, and the goal is to predict which labels are contained in each node. 

We use essentially the same reference architecture as for point cloud segmentation described in \secLabel \ref{sec:PCGCN}, but we predict multiple labels. The number of filters of the first and last layers are changed to adapt to this task. Instead of constructing edges by means of $k$-NN, we use the edges provided by PPI directly. If we were to construct edges dynamically, there is a chance to lose the rich information provided by the initial edges.

\subsection{Experimental Setup}
Following common procedure, we use the micro-average F1 (m-F1) score as the evaluation metric. 
For each graph, the F1 score is computed as follows:
\begin{equation}
\text{F1-score} = 2 \times \frac{(precision  \times recall)}{(precision+recall)},
\end{equation}
where $precision = \frac{TP'}{P'}$, $recall = \frac{TP'}{T'}$, $TP'$ is the total number of true positive points of all the classes, $T'$ is the number of ground truth positive points, and $P'$ is the number of predicted positive points. 
We find the best model, \ie the one with the highest accuracy (m-F1 score) on the validation set in the training phase, and then calculate the m-F1 score across all the graphs in the test set.

We show the performance and GPU memory usage of our proposed \emph{MRGCN} and compare them with other graph convolutions, \eg EdgeConv \cite{wang2018dynamic}, GATConv \cite{velivckovic2017graph}, SemiGCN \cite{kipf2016semi} and GINConv \cite{xu2018powerful}.
We conduct an extensive ablation study on the number of filters and the number of GCN layers to show their effect in the backbone network. 
Our ablation study also includes experiments to show the importance of residual graph connections and dense graph connections in our DeepGCN framework. To ensure a fair comparison, all networks in our ablation study share the same architecture.
Finally, we compare our best models to several state-of-the-art methods for this task. 

\subsection{Implementation}
For this biological network node classification task, we implement all our models based on PyTorch Geometric \cite{Fey/Lenssen/2019}. 
We use the Adam optimizer with the same initial learning rate $0.0002$ and learning rate schedule with learning rate decay of $80\%$ every $2,000$ gradient decent steps for all the experiments. 
The networks are trained with one NVIDIA Tesla V100 GPU with a batch size of $1$. Dropout with a rate of $0.2$ is used at the first and second MLP layers of the prediction block. 
For fair comparison, we do not use any data augmentation or post processing techniques. 
Our models are trained end-to-end from scratch.

\subsection{Results}
We study the effect of residual and dense graph connections on multi-label node classification performance. We also investigate the influence of different parameters, \eg the number of filters $(32, 64, 128, 256)$ and layers $(3, 7, 14, 28, 56, 112)$. To show the generality of our framework, we also apply the proposed residual connections to multiple GCN variants.

\mysection{Effect of graph connections}
Results in \tblLabel \ref{tb: ppi-deep} show that both residual and dense graph connections help train deeper networks. When the network is shallow, models with graph connections achieve similar performance as models without them. However, as the network grows deeper, the performance of models without graph connections drops dramatically, while the performance of models with graph connections is stable or even improves further. For example, when the number of filters is 32 and the depth is 112, the performance of \emph{ResMRGCN-112} is nearly 37.66\% higher than \emph{PlainMRGCN-112} in terms of the m-F1 score. We note that \emph{DenseMRGCN} achieves slightly better performance than \emph{ResMRGCN} with the same network depth and width. 

\mysection{Effect of network depth}
Results in \tblLabel \ref{tb: ppi-deep} show that increasing the number of layers improves network performance if residual  or dense graph connections are used. Although \emph{ResMRGCN} has a slight performance drop when the number of layers reaches $112$, the m-F1 score is still much higher than the corresponding \emph{PlainMRGCN}. The performance of \emph{DenseMRGCN} increases reliably as the network grows deeper; however, \emph{DenseMRGCN} consumes more memory than \emph{ResMRGCN} due to concatenations of feature maps. Due to this memory issue, we are unable to train some models and denote them with '-' in \tblLabel \ref{tb: ppi-deep}. Meanwhile, \emph{PlainMRGCN}, which has no graph connections, only enjoys a slight performance gain as the network depth increases from $3$ to $14$. For depths beyond $14$ layers, the performance drops significantly. Clearly, using graph connections improves performance, especially for deeper networks where it becomes essential.

\mysection{Effect of network width} 
Results of each row in \tblLabel \ref{tb: ppi-deep}  show that increasing the number of filters can consistently increase performance. A higher number of filters can also help convergence for deeper networks. However, a large number of filters is very memory consuming. Hence, we only consider networks with up to $256$ filters in our experiments.

\mysection{Effect of GCN variants}
\tblLabel \ref{tb: ppi-convs} shows the effect of using different GCN operators with different model depths. Residual graph connections and the GCN operators are the only difference when the number of layers is kept the same. The results clearly show that residual graph connections in deep networks can help different GCN operators achieve better performance than \emph{PlainGCN}. Interestingly, when the network grows deeper, the performance of \emph{PlainSemiGCN}, \emph{PlainGAT}, and \emph{PlainGIN} decreases dramatically; meanwhile, \emph{PlainEdgeConv} and \emph{PlainMRGCN} only observe a relatively small performance drop. In comparison, our proposed \emph{MRGCN} operator achieves the best performance among all models.

\begin{figure}[!t]
    \centering
    \begin{tabular}{cc}
    \includegraphics[trim=5mm 0mm 0mm 0mm, width=1\columnwidth]{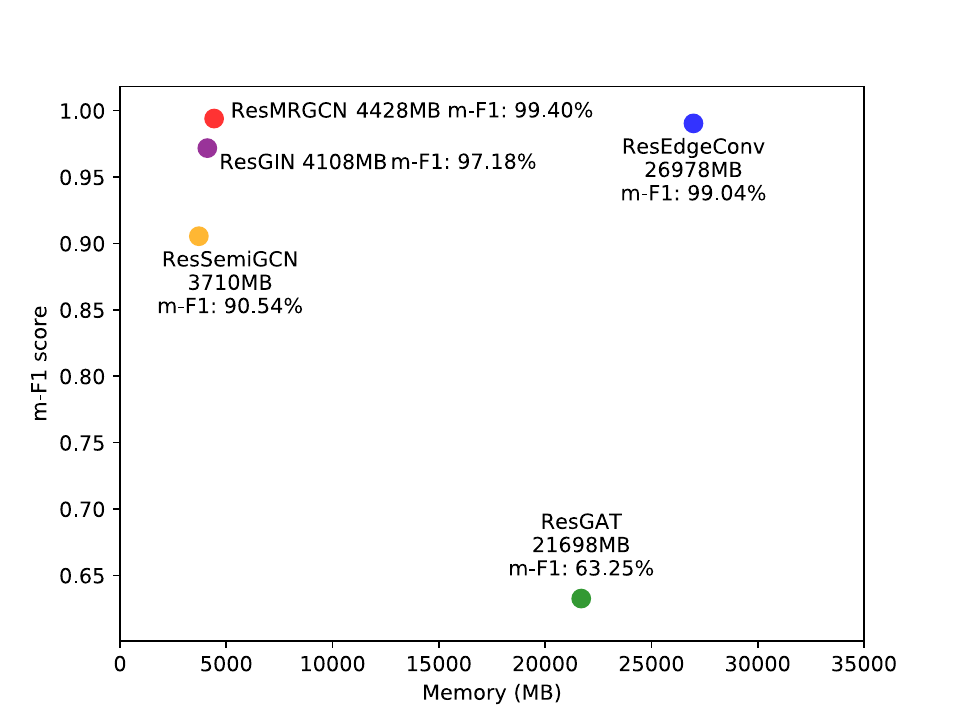}
    \end{tabular}
    \caption{\textbf{Memory usage for different GCNs on PPI node classification.} We keep the same parameters for different models and compare their m-F1 score with their total memory usage. All models in the comparison have 256 filters per layer and 56 layers. We notice that our model \emph{ResMRGCN} uses approximately 1/6 of the total memory used by \emph{ResEdgeConv} and gives a better m-F1 score.}
    \label{fig: ppi-memory}
\end{figure}

\mysection{Memory usage of GCN variants}
In \figLabel \ref{fig: ppi-memory}, we compare the total memory usage and performance of different GCN operators. All these models share the the same architecture except for the GCN operations. They all use residual graph connections with  $56$ layers and $256$ filters. We implement all the models with PyTorch Geometric and train each using one NVIDIA Tesla V100. The GPU memory usage is measured when the memory usage is stable. Our proposed \emph{ResMRGCN} achieves the best performance, while only using around $15$\% GPU memory compared to \emph{ResEdgeConv}.  

\mysection{Comparison to state-of-the-art} 
Finally, we compare our \emph{DenseMRGCN-14} and \emph{ResMRGCN-28} to several state-of-the-art baselines in \tblLabel \ref{tb: ppi-sota}. Results clearly show the effectiveness of deeper models with residual and dense graph connections. \emph{DenseMRGCN-14} and \emph{ResMRGCN-28} outperform the previous state-of-the-art Cluster-GCN \cite{chiang2019cluster} by 0.07\% and 0.05\% respectively. It is worth mentioning that a total of ten models in \tblLabel \ref{tb: ppi-deep} surpass Cluster-GCN.

\begin{table}[!htb]
\centering
\setlength{\tabcolsep}{5pt} 
\begin{tabular}{l|cccc}

\toprule
\textbf{Number of filters} & 32 & 64 & 128 & 256 \\ 
\midrule
\emph{PlainMRGCN-3} & 95.84 & 97.60 & 98.58 & 99.13 \\
\emph{PlainMRGCN-7} & 97.35 & 98.69 & 99.22 & \textbf{99.38}\\ 
\emph{PlainMRGCN-14} & 97.55 & 99.02 & 99.31 & 99.34\\ 
\emph{PlainMRGCN-28} & 98.09 & 99.00 & 99.02 & 99.31\\
\emph{PlainMRGCN-56} & 92.70 & 97.43 & 97.31 & 97.61\\ 
\emph{PlainMRGCN-112} & 60.75 & 71.97 & 89.69 & 91.50\\ 
\midrule
\emph{ResMRGCN-3} & 96.04 & 97.60 & 98.53 & 99.09\\
\emph{ResMRGCN-7} & 97.00 & 98.43 & 99.19 & 99.30\\ 
\emph{ResMRGCN-14} & 97.75 & 98.88 & 99.26 & \textbf{99.38}\\
\emph{ResMRGCN-28} & 98.50 & 99.16 & 99.29 & \textbf{99.41}\\
\emph{ResMRGCN-56} & 98.62 & 99.27 & \textbf{99.36} & \textbf{99.40}\\ 
\emph{ResMRGCN-112} & 98.41 & 99.34 & \textbf{99.38} & \textbf{99.39}\\ 
\midrule
\emph{DenseMRGCN-3} & 95.96 & 97.85 & 98.66 & 99.11\\ 
\emph{DenseMRGCN-7} & 97.87 & 98.47 & 99.31 & \textbf{99.36}\\ 
\emph{DenseMRGCN-14} & 98.93 & 99.00 & 99.01 & \textbf{99.43}\\ 
\emph{DenseMRGCN-28} & 99.16 & 99.29 & \textbf{99.42} & -\\ 
\emph{DenseMRGCN-56} & 99.22 & - & - & -\\ 
\bottomrule

\end{tabular}
\vspace{3pt}
\caption{\textbf{Ablation study on graph connections, network width, and network depth}. The m-F1 score is used as the evaluation metric (in \%). We find that residual and dense connections can help deep networks converge much better compared to the same model without any connections. Also, network width is positively correlated with network performance. Note that '-' denotes models that are not applicable due to memory limitations. \textbf{Bold} highlights models that outperform all state-of-the-art baselines for this multi-label PPI node classification problem.
}
\label{tb: ppi-deep}
\end{table}

\begin{table}[!htb]
\centering
\setlength{\tabcolsep}{5pt} 
\begin{tabular}{l|ccccc}
\toprule
\textbf{Number of layers} & 3 & 7 & 14 & 28 & 56\\ 
\midrule
\emph{PlainSemiGCN} & 97.82 & 90.40 & 80.55 & 41.00 & 50.75\\
\emph{ResSemiGCN} & 97.88 & 95.05 & 93.50 & 90.60 & 90.54\\ 
\midrule
\emph{PlainGAT} & 98.52 & 80.92 & 56.88 & 42.40 & 48.95\\
\emph{ResGAT} & 98.63 & 97.86 & 98.99 & 99.06 & 63.25\\ 
\midrule
\emph{PlainGIN} & 97.86 & 57.78 & 40.79 & 35.82 & 0.26\\ 
\emph{ResGIN} & 97.80 & 96.44 & 98.22 & 97.44 & 97.18\\ 
\midrule
\emph{PlainEdgeConv} & 99.16 & 99.27 & 99.30 & 99.33 & 98.99\\ 
\emph{ResEdgeConv} & 99.03 & 99.19 & 99.26 & 99.30 & 99.04\\ 
\midrule
\emph{PlainMRGCN} & 99.13 & \textbf{99.38} & 99.34 & 99.31 & 97.61\\
\textbf{\emph{ResMRGCN}} & 99.09 & 99.30 & \textbf{99.38} & \textbf{99.41} & \textbf{99.40}\\
\bottomrule

\end{tabular}
\vspace{3pt}
\caption{\textbf{Ablation study on network depth and GCN variants.} m-F1 score is used as the evaluation metric (in \%). We set the number of filters per layer in the backbone network to $256$ and vary the number of layers. Residual graph connections can generally help different GCN operators achieve better performance than \emph{PlainGCN} when the network grows deep. \textbf{Bold} highlights models that outperform all state-of-the-art baselines for this multi-label PPI node classification problem.
}
\label{tb: ppi-convs}
\end{table}

\begin{table}[!htb]
\centering
\setlength{\tabcolsep}{5pt} 
\begin{tabular}{l|c}
\toprule
\textbf{Model} & \textbf{m-F1 score (\%)} \\ 
\midrule
GraphSAGE\cite{hamilton2017inductive} & 61.20 \\ 
GATConv \cite{velivckovic2017graph} & 97.30 \\ 
VR-GCN \cite{chen2017vrgcn} & 97.80 \\ 
GaAN \cite{zhang2018gaan} & 98.71 \\ 
GeniePath \cite{liu2019geniepath} & 98.50 \\ 
Cluster-GCN \cite{chiang2019cluster}& 99.36 \\ 
\midrule
\textbf{\emph{ResMRGCN-28} (\textit{Ours})} & \textbf{99.41} \\ 
\textbf{\emph{DenseMRGCN-14} (\textit{Ours})} & \textbf{99.43} \\ 
\bottomrule
\end{tabular}
\vspace{3pt}
\caption{\textbf{Comparison of \emph{DenseMRGCN-14} with state-of-the-art on PPI node classification}. We follow convention and compare models based on the m-F1 score. Our model \emph{DenseMRGCN-14}, which has 14 MRGCN layers, 256 filters in the first layer, and dense graph connections outperforms all baselines. Our  \emph{ResMRGCN-28}, which has 28 MRGCN layers, 256 filters per layer, and residual graph connections also outperforms previous state-of-the-art.}
\label{tb: ppi-sota}
\end{table}

\section{Conclusion}\label{sec:conclusion}
This work shows how proven concepts from CNNs (\ie residual/dense connections and dilated convolutions) can be transferred to GCNs in order to make GCNs go as deep as CNNs. Adding skip connections and dilated convolutions to GCNs alleviates the training difficulty, which was impeding GCNs to go deeper and thus impeding further progress. We also encourage readers to read the recent work on benchmarking GNNs \cite{dwivedi2020benchmarking} to further understand how different techniques can aid in training GNNs.

A large number of experiments on semantic segmentation and part segmentation of 3D point clouds, as well as node classification on biological graphs show the benefit of deeper architectures, as they achieve state-of-the-art performance. We also show that our approach generalizes across several GCN operators. 
For the point cloud tasks, we achieve the best results using EdgeConv \cite{wang2018understanding} as GCN operators for our backbone networks. Moreover, we find that dilated graph convolutions help to gain a larger receptive field without loss of resolution. Even with a small number of nearest neighbors, DeepGCNs can achieve high performance on point cloud semantic segmentation. \emph{ResGCN-112} and \emph{ResGCN-56} perform very well on this task, although they only use $4$ and $8$ nearest neighbors respectively compared to $16$ for \emph{ResGCN-28}. For the biological graph task, we achieve the best results using the MRGCN operator, which we propose as a novel memory-efficient alternative to EdgeConv. \RR{We successfully trained \emph{ResMRGCN-112} and \emph{DenseMRGCN-56}; both networks converged very well and achieved very promising results on the PPI dataset.}

\section{Future Work}\label{sec:future}
Our results show that after solving the vanishing gradient problem plaguing deep GCNs, we can either make GCNs deeper or wider to get better performance. We expect GCNs to become a powerful tool for processing graph-structured data in computer vision, natural language processing, and data mining. We show successful cases for adapting concepts from CNNs to GCNs (\ie skip connections and dilated convolutions). In the future, it will be worthwhile to explore how to transfer other operators (\eg deformable convolutions \cite{dai2017deformable}), other architectures (\eg feature pyramid architectures \cite{zhao2017pyramid}), \etc. It will also be interesting to study different distance measures to compute dilated $k$-NN, constructing graphs with different $k$ at each layer, better dilation rate schedules \cite{chen2017rethinking, wang2018understanding} for GCNs, and combining residual and dense connections.
We also point out that, for the specific task of point cloud semantic segmentation, the common approach of processing the data in $1m\times 1m$ columns is sub-optimal for graph representation. A more suitable sampling approach should lead to further performance gains on this task. For the task of node classification, the existing datasets are relatively small. We expect that experimenting on larger datasets will further unleash the full potential of \emph{DeepGCNs}.
As evidence, a recent work \cite{li2020deepergcn} shows that deep residual GNNs yield promising results on large-scale graph datasets \cite{hu2020open}.


%



\ifCLASSOPTIONcompsoc
  \section*{Acknowledgments}
\else
  \section*{Acknowledgment}
\fi

The authors thank Adel Bibi and Hani Itani for their help with the project. This work was supported by the King Abdullah University of Science and Technology (KAUST) Office of Sponsored Research through the Visual Computing Center (VCC) funding. We also thank the editors and reviewers for their constructive suggestions.

\ifCLASSOPTIONcaptionsoff
  \newpage
\fi



%
\bibliographystyle{IEEEtran}
\bibliography{references}
%

\begin{IEEEbiography}[{\includegraphics[width=1in,height=1.25in,clip,keepaspectratio]{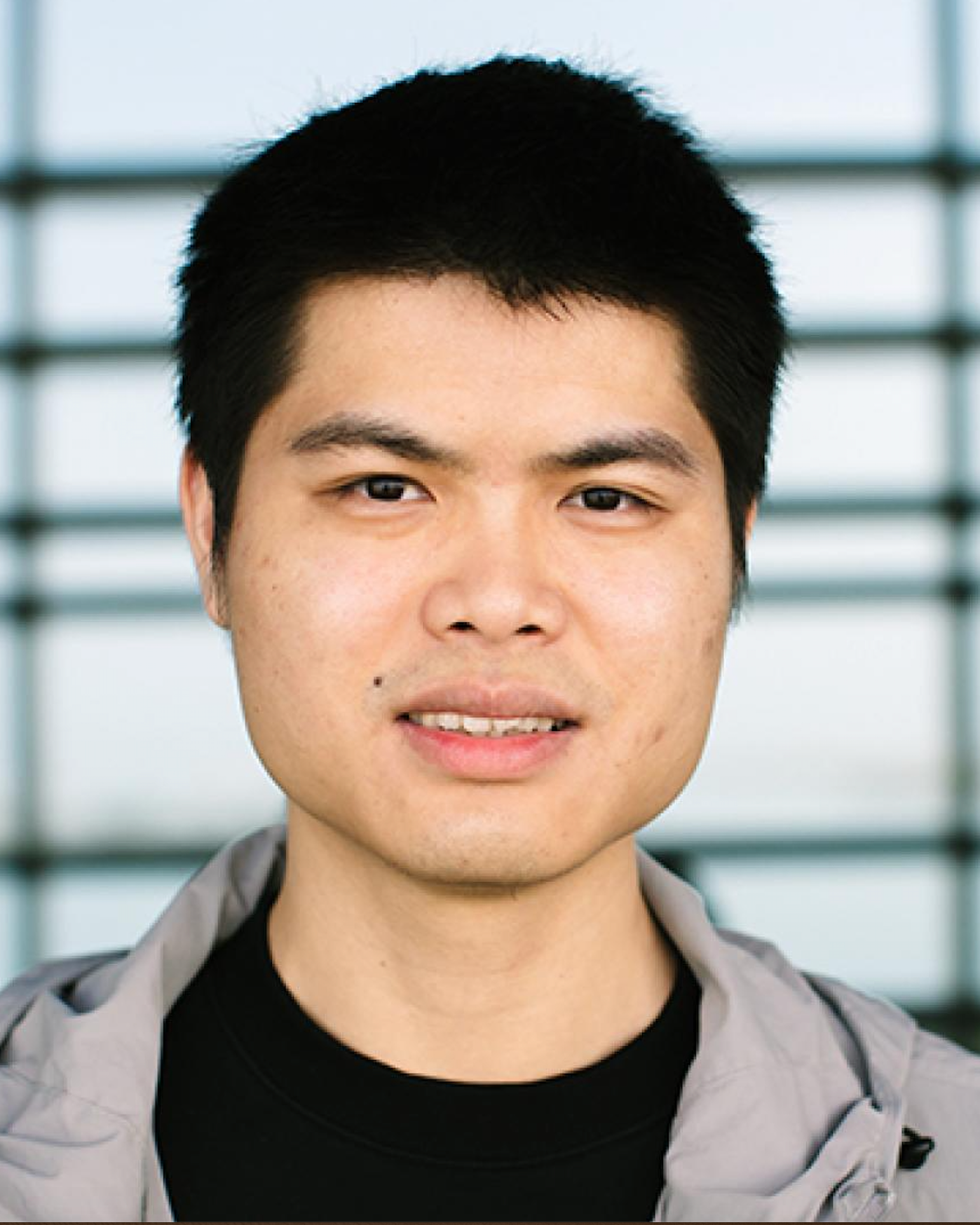}}]{Guohao Li} obtained his BEng degree in Communication Engineering from Harbin Institute of Technology in 2015. In 2018, he received his Master Degree in Communication and Information Systems from Chinese Academy of Science. He was a research intern at SenseTime and Intel ISL. He is currently a CS PhD student at King Abdullah University of Science and Technology. His primary research interests are Computer Vision, Robotics and Deep Learning.
\end{IEEEbiography}

\begin{IEEEbiography}[{\includegraphics[width=1in,height=1.25in,clip,keepaspectratio]{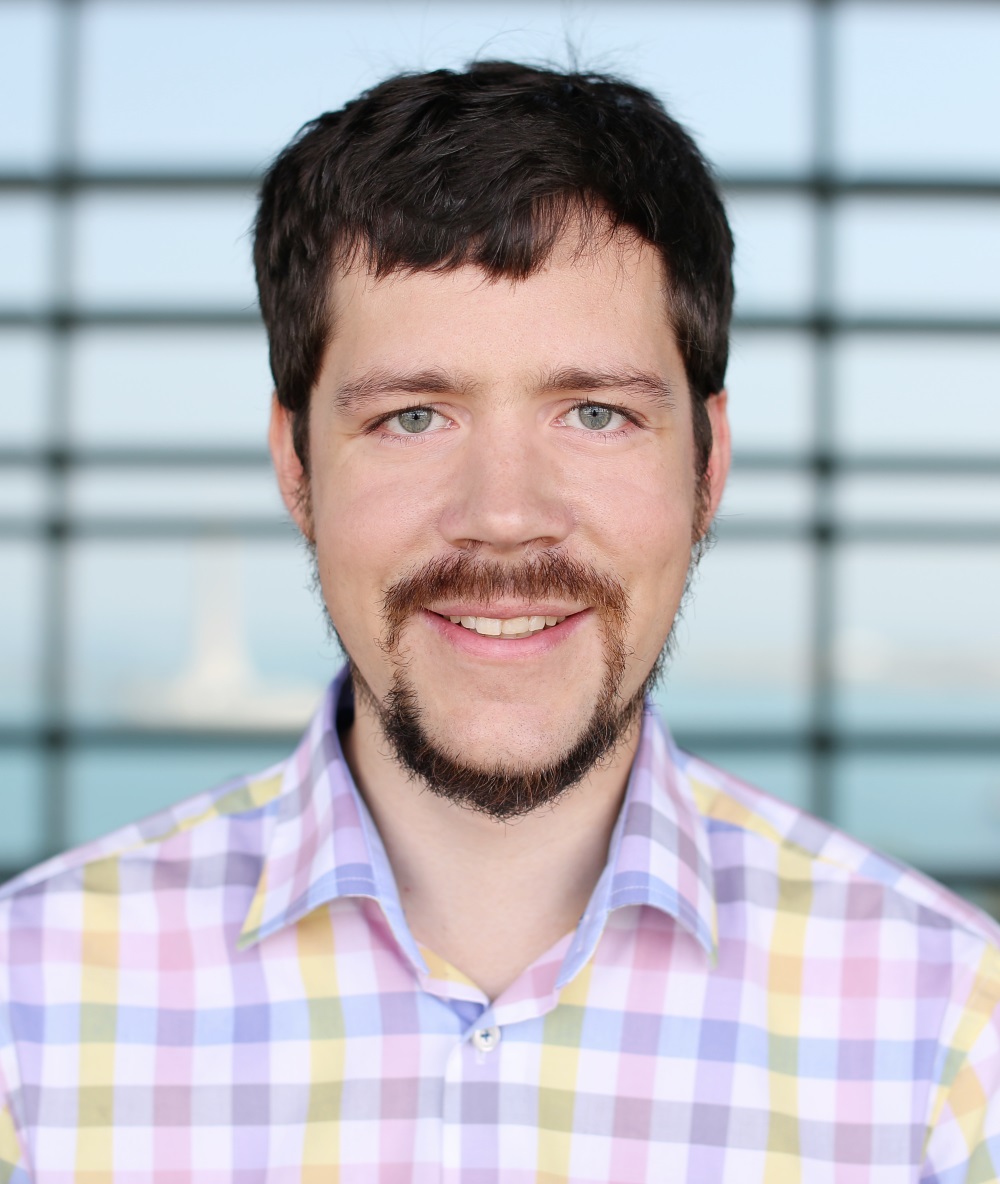}}]{Matthias M{\"u}ller} received his PhD in computer vision from KAUST in 2019. He now works as a research scientist at the Intelligent Systems Lab at Intel. His research interests lie in the fields of computer vision, robotics and machine learning where he has contributed to more than 10 publications in top tier conferences and journals. Matthias was recognized as an outstanding reviewer for CVPR'18 and won the best paper award at the ECCV'18 workshop UAVision.
\end{IEEEbiography}

\begin{IEEEbiography}[{\includegraphics[width=1in,height=1.25in,clip,keepaspectratio]{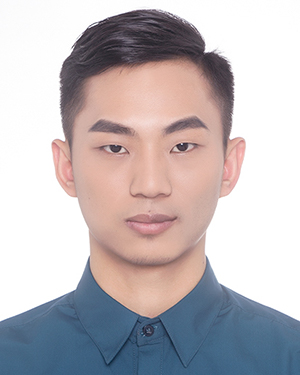}}]{Guocheng Qian}
received the BEng degree with first class honors from Xi'an Jiaotong University, China in 2018.  He is working towards the MSc degree currently in the Department of Computer Science at King Abdullah University of Science and Technology.  His research interests include computer vision, computational photography and neural architecture search. 
\end{IEEEbiography}

\begin{IEEEbiography}[{\includegraphics[width=1in,height=1.25in,clip,keepaspectratio]{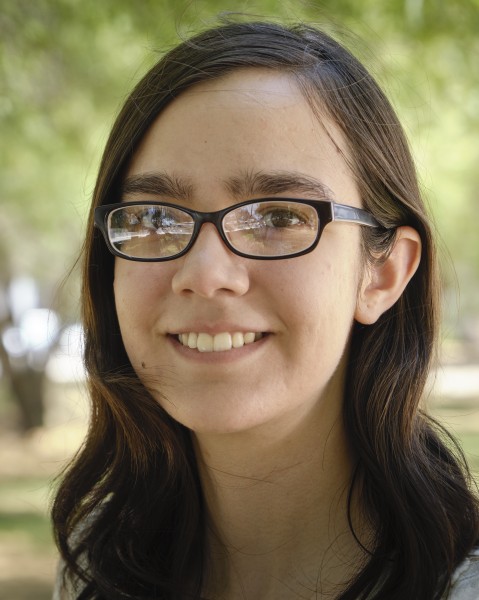}}]{Itzel C. Delgadillo}
received the Bachelor in Artificial Intelligence from the Panamerican University, Mexico in 2019. She is currently a Visiting Student Research Intern in King Abdullah University of Science and Technology. Her research interests include computer vision and neural architecture search.
\end{IEEEbiography}

\begin{IEEEbiography}[{\includegraphics[width=1in,height=1.25in,clip,keepaspectratio]{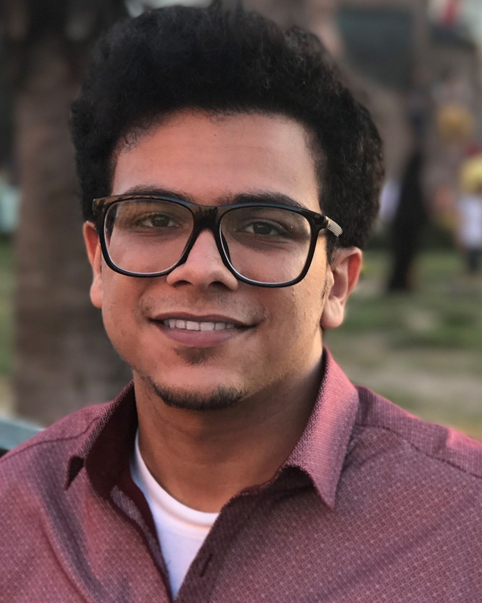}}]{Abdulellah Abualshour}
received the BS degree in computer science from Rutgers, The State University of New Jersey in 2018. He was a recipient of the KAUST Gifted Student Program (KGSP) scholarship award. He is currently an MS student and a member of the Image and Video Understanding Lab (IVUL) at the Visual Computing Center (VCC) at King Abdullah University of Science and Technology (KAUST). His research interests include computer vision and deep learning.
\end{IEEEbiography}

\begin{IEEEbiography}[{\includegraphics[width=1in,height=1.25in,clip,keepaspectratio]{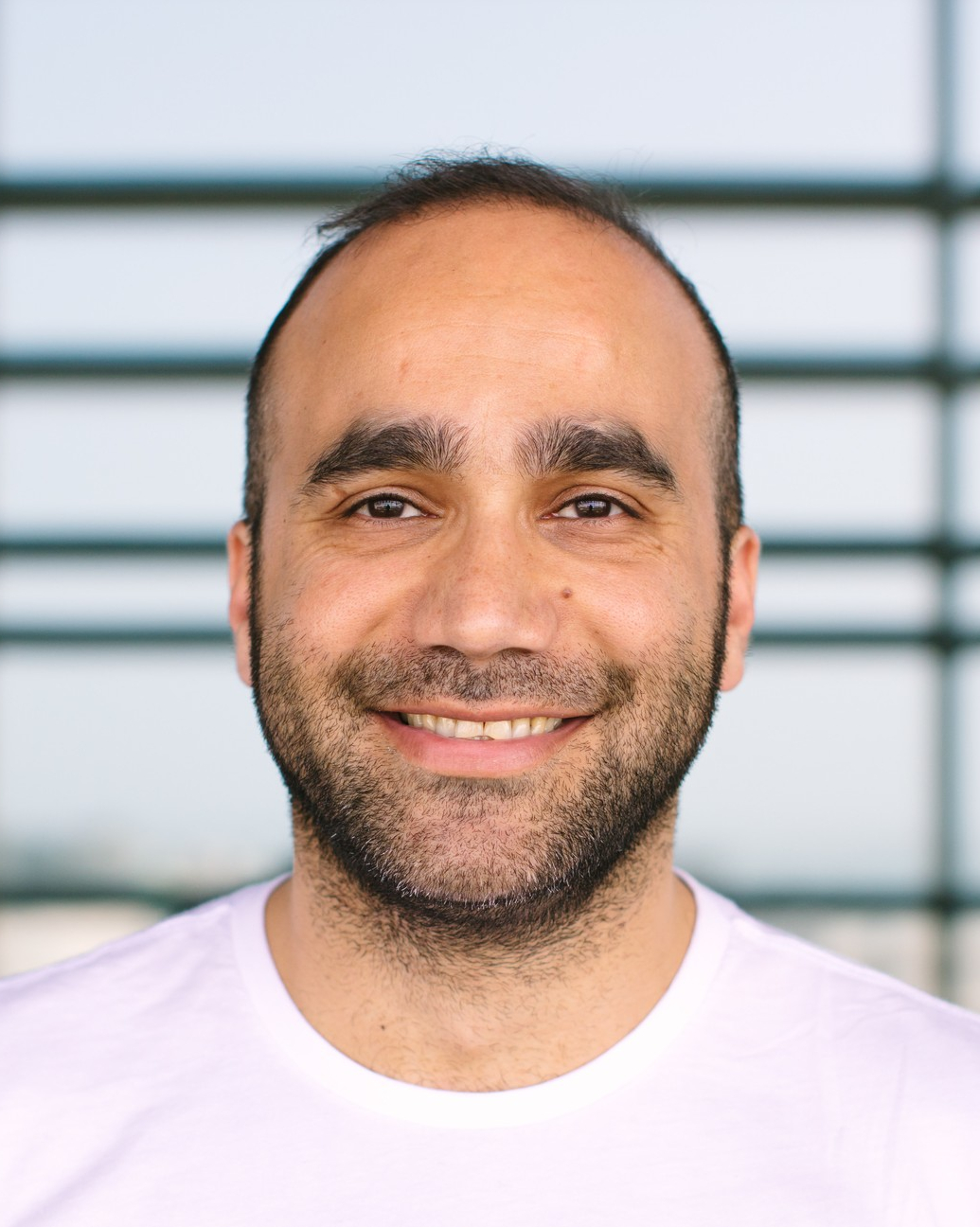}}]{Ali Thabet}
is a Research Scientist at the Visual Computing Center (VCC) in King Abdullah University of Science and Technology (KAUST), working in the Image and Video Understanding Laboratory (IVUL). His research focuses on problems related to 3D computer vision. In general, he’s interested in algorithmic applications of machine learning, and deep learning specifically, to understand the 3D world with the help of sensors like RGB-D cameras, LiDAR, and others. Also, Ali is interested in applying deep learning to image reconstruction, 3D object detection and generation, and autonomous vehicles.
\end{IEEEbiography}

\begin{IEEEbiography}[{\includegraphics[width=1in,height=1.25in,clip,keepaspectratio]{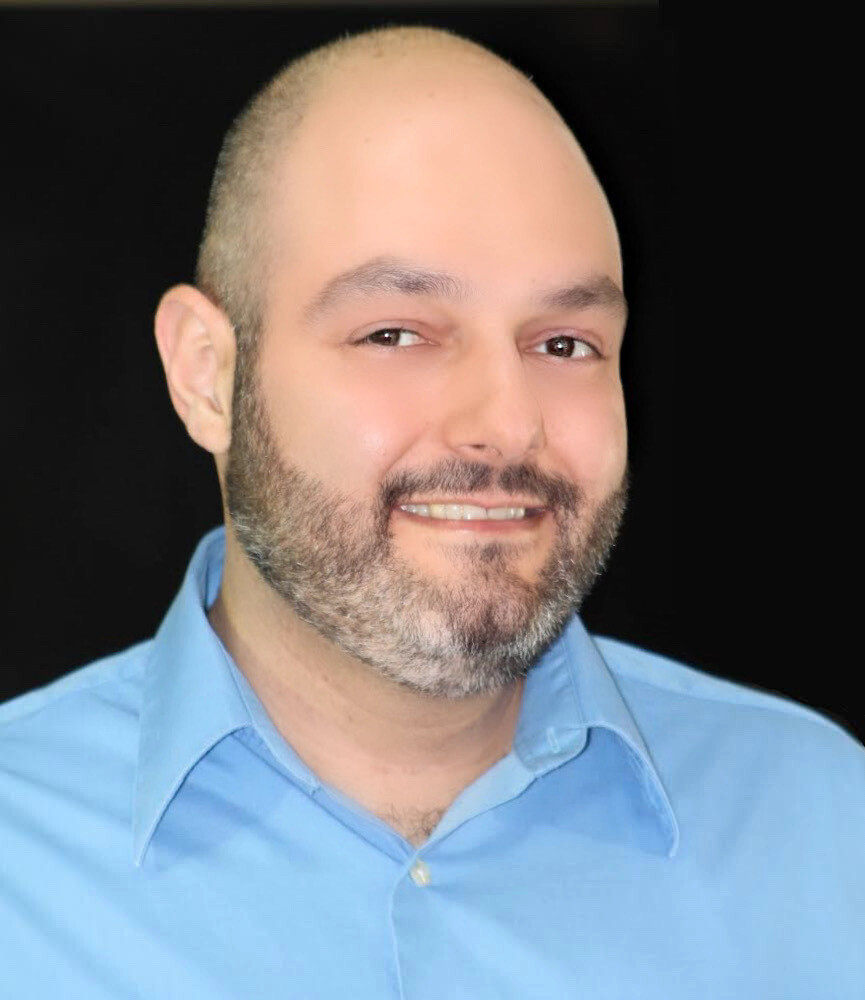}}]{Bernard Ghanem} is currently an Associate Professor in the CEMSE division, a theme leader at the Visual Computing Center (VCC), and the Interim Lead of the AI Initiative at King Abdullah University of Science and Technology (KAUST). His research interests lie in computer vision and machine learning with emphasis on topics in video understanding, 3D recognition, and theoretical foundations of deep learning. He received his Bachelor’s degree from the American University of Beirut (AUB) in 2005 and his MS/PhD from the University of Illinois at Urbana-Champaign (UIUC) in 2010. His work has received several awards and honors, including four Best Paper Awards for workshops in CVPR 2013\&2019 and ECCV 2018\&2020, a Google Faculty Research Award in 2015 (1st in MENA for Machine Perception), and a Abdul Hameed Shoman Arab Researchers Award for Big Data and Machine Learning in 2020. He has co-authored more than 120 peer reviewed conference and journal papers in his field as well as three issued patents. He serves as an Associate Editor for IEEE Transactions on Pattern Analysis and Machine Intelligence (TPAMI) and has served as Area Chair (AC) for CVPR 2018/2021, ICCV 2019/2021, ICLR 2021, and AAAI 2021.
\end{IEEEbiography}




\newpage

\section{Qualitative Results for DeepGCNs} \label{appendix:qualitative_results}
\begin{figure*}[!h]
    \centering
    \includegraphics[height=\textheight]{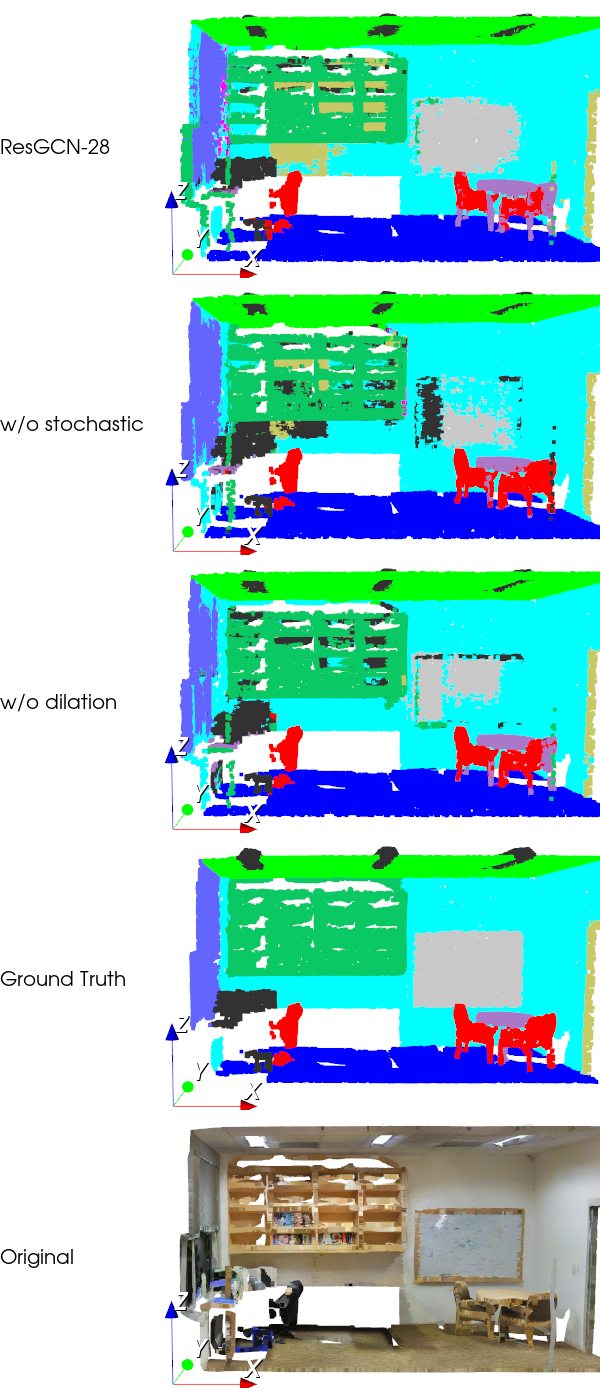}
    \caption{\textbf{Qualitative Results for S3DIS Semantic Segmentation}. We show the importance of stochastic dilated convolutions.}
    \label{fig:qualitative_dilation}
\end{figure*}

\begin{figure*}[!h]
    \centering
    \includegraphics[height=\textheight]{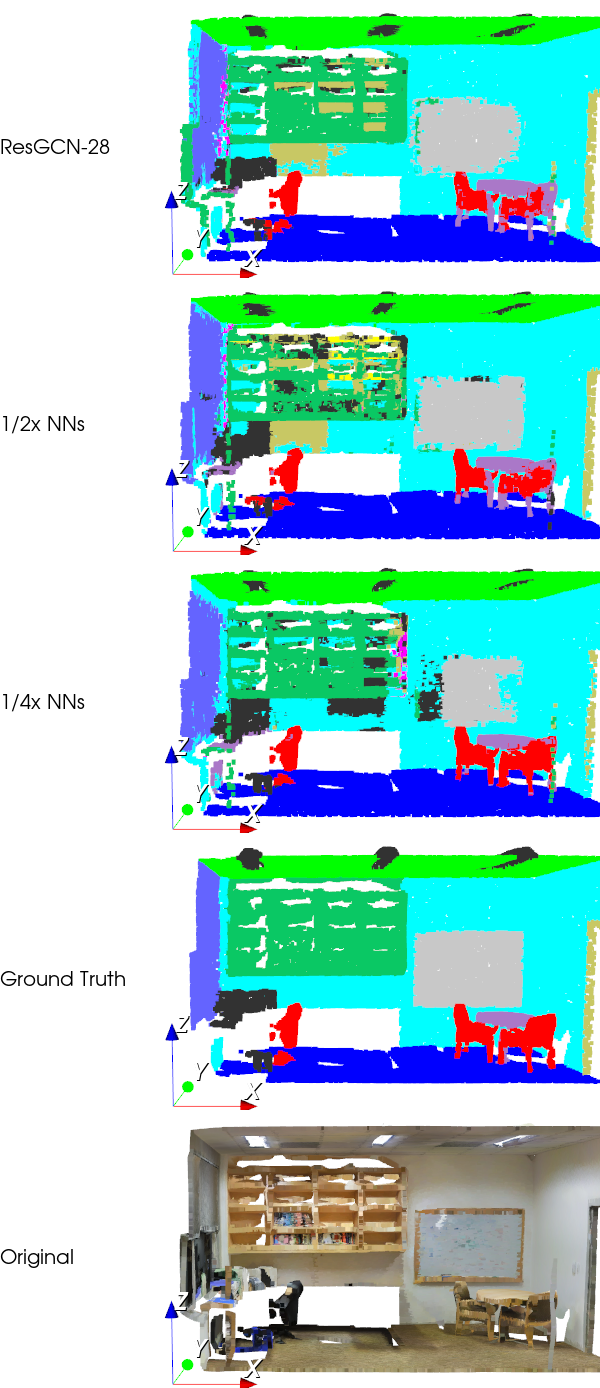}
    \caption{\textbf{Qualitative Results for S3DIS Semantic Segmentation}. We show the importance of the number of nearest neighbors used in the convolutions.}
    \label{fig:qualitative_nearest_neighbors}
\end{figure*}

\begin{figure*}[!h]
    \centering
    \includegraphics[height=\textheight]{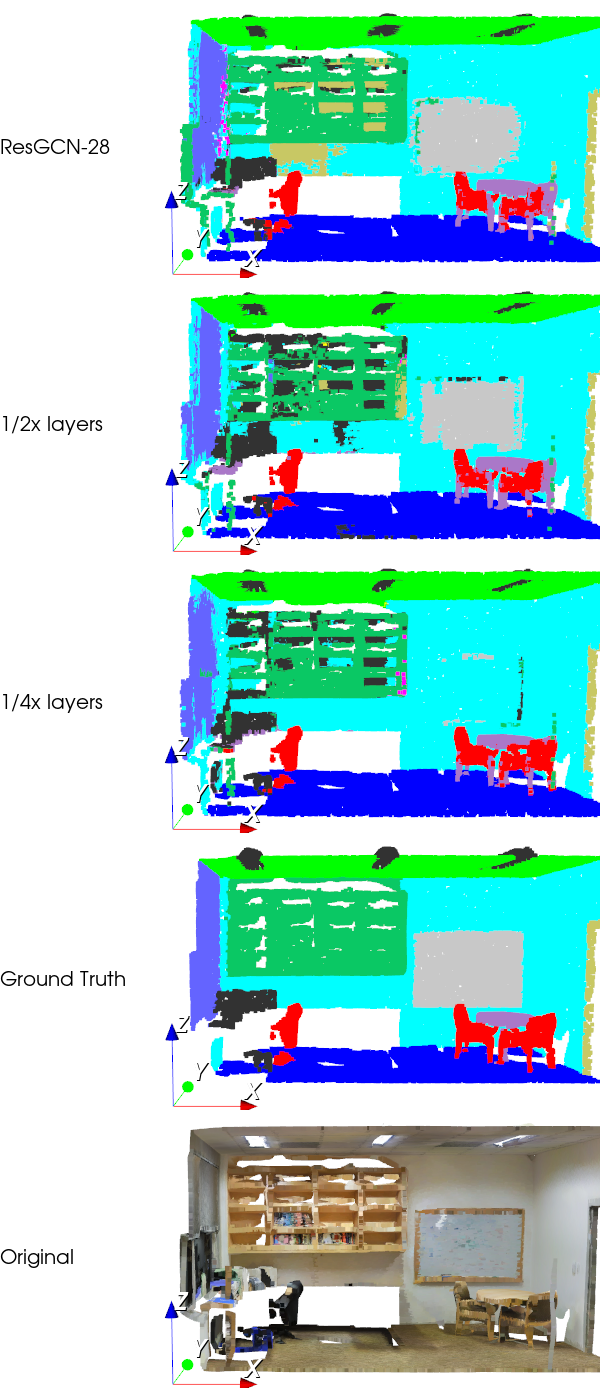}
    \caption{\textbf{Qualitative Results for S3DIS Semantic Segmentation}. We show the importance of network depth (number of layers).}
    \label{fig:qualitative_layers}
\end{figure*}

\begin{figure*}[!h]
    \centering
    \includegraphics[height=\textheight]{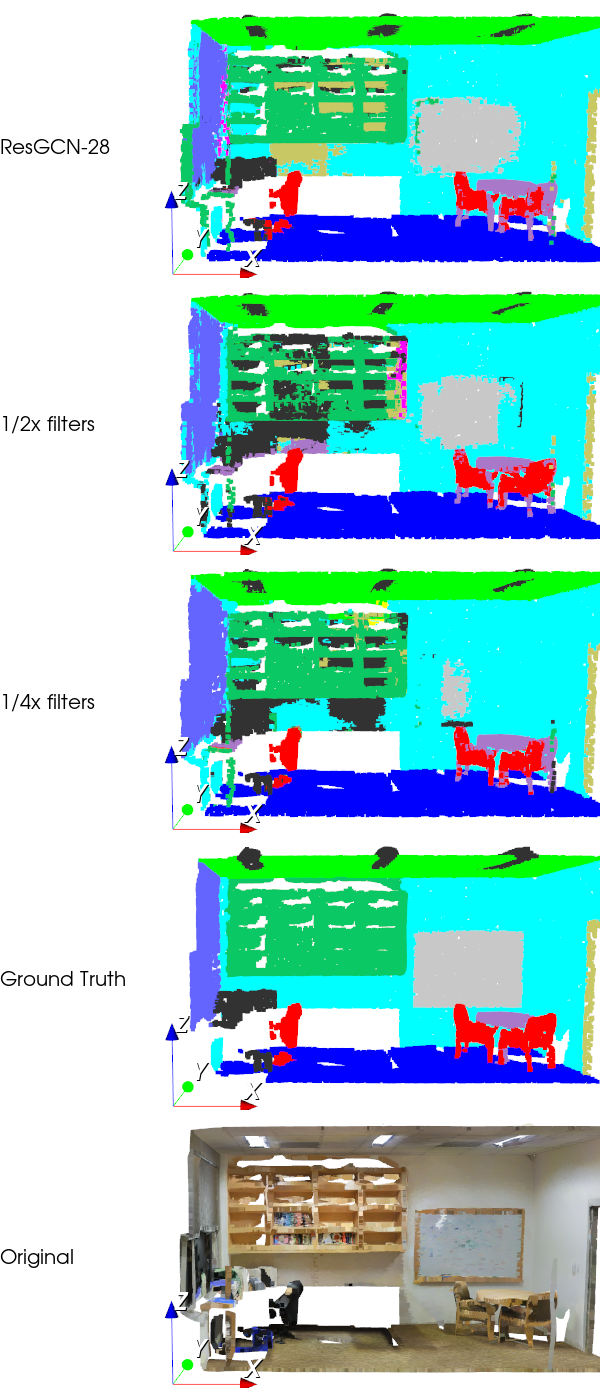}
    \caption{\textbf{Qualitative Results for S3DIS Semantic Segmentation}. We show the importance of network width (number of filters per layer).}
    \label{fig:qualitative_filters}
\end{figure*}

\begin{figure*}[!h]
    \centering
    \includegraphics[height=\textheight]{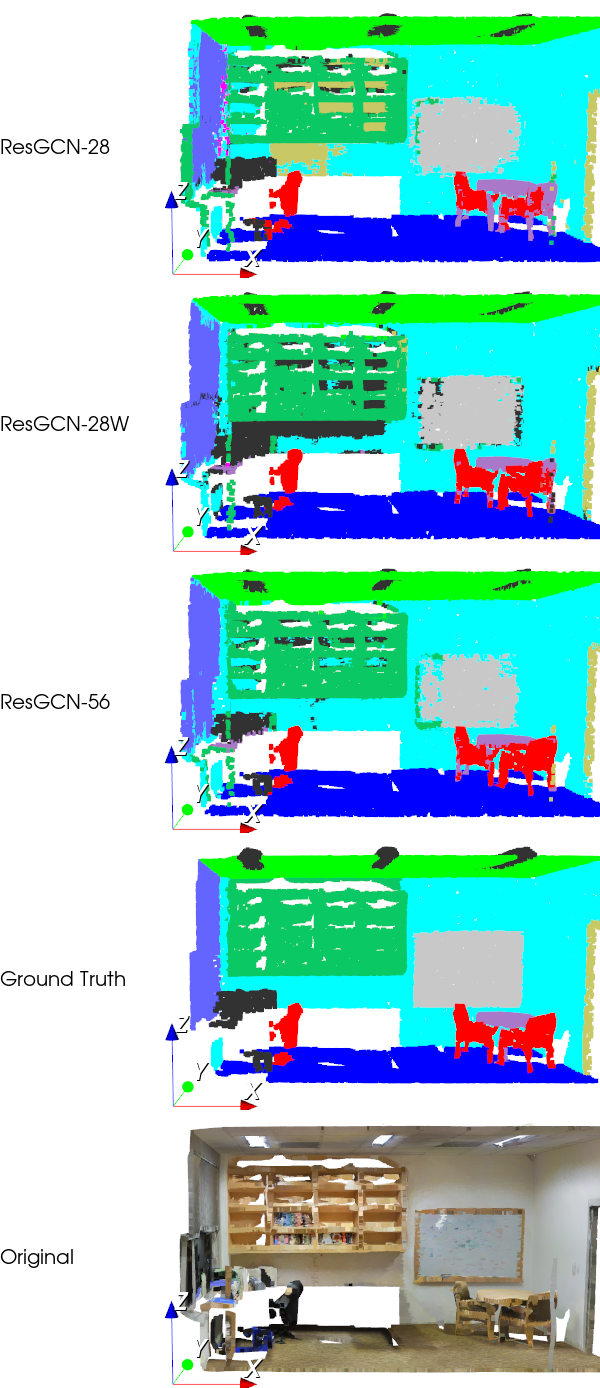}
    \caption{\textbf{Qualitative Results for S3DIS Semantic Segmentation}. We show the benefit of a wider and deeper network even with only half the number of nearest neighbors.}
    \label{fig:qualitative_56_28W}
\end{figure*}

\begin{figure*}[!h]
    \centering
    \includegraphics[page=1,trim = 0mm 0mm 200mm 300mm, clip, width=3.55in]{figures/PlainGCN_28_vs_ResGCN_28.pdf}
    \includegraphics[page=3,trim = 0mm 0mm 200mm 300mm, clip, width=3.55in]{figures/PlainGCN_28_vs_ResGCN_28.pdf}
    \includegraphics[page=5,trim = 0mm 0mm 200mm 600mm, clip, width=3.55in]{figures/PlainGCN_28_vs_ResGCN_28.pdf}
    \includegraphics[page=2,trim = 0mm 0mm 200mm 600mm, clip, width=3.55in]{figures/PlainGCN_28_vs_ResGCN_28.pdf}
    \includegraphics[page=4,trim = 0mm 0mm 200mm 600mm, clip, width=3.55in]{figures/PlainGCN_28_vs_ResGCN_28.pdf}
    \includegraphics[page=6,trim = 0mm 0mm 200mm 600mm, clip, width=3.55in]{figures/PlainGCN_28_vs_ResGCN_28.pdf}
    \includegraphics[page=7,trim = 0mm 0mm 200mm 600mm, clip, width=3.55in]{figures/PlainGCN_28_vs_ResGCN_28.pdf}
    \includegraphics[page=8,trim = 0mm 0mm 200mm 600mm, clip, width=3.55in]{figures/PlainGCN_28_vs_ResGCN_28.pdf}
    \includegraphics[page=9,trim = 0mm 0mm 200mm 600mm, clip, width=3.55in]{figures/PlainGCN_28_vs_ResGCN_28.pdf}
    \includegraphics[page=10,trim = 0mm 0mm 200mm 600mm, clip, width=3.55in]{figures/PlainGCN_28_vs_ResGCN_28.pdf}
    \includegraphics[page=11,trim = 0mm 0mm 200mm 600mm, clip, width=3.55in]{figures/PlainGCN_28_vs_ResGCN_28.pdf}
    \includegraphics[page=12,trim = 0mm 0mm 200mm 600mm, clip, width=3.55in]{figures/PlainGCN_28_vs_ResGCN_28.pdf}
    \includegraphics[page=13,trim = 0mm 0mm 200mm 600mm, clip, width=3.55in]{figures/PlainGCN_28_vs_ResGCN_28.pdf}
    \includegraphics[page=14,trim = 0mm 0mm 200mm 600mm, clip, width=3.55in]{figures/PlainGCN_28_vs_ResGCN_28.pdf}
    \includegraphics[page=15,trim = 0mm 0mm 200mm 600mm, clip, width=3.55in]{figures/PlainGCN_28_vs_ResGCN_28.pdf}
    \includegraphics[page=16,trim = 0mm 0mm 200mm 600mm, clip, width=3.55in]{figures/PlainGCN_28_vs_ResGCN_28.pdf}
    \includegraphics[page=17,trim = 0mm 0mm 200mm 600mm, clip, width=3.55in]{figures/PlainGCN_28_vs_ResGCN_28.pdf}
    \includegraphics[page=1, trim = 0mm 0mm 200mm 600mm, clip, width=3.55in]{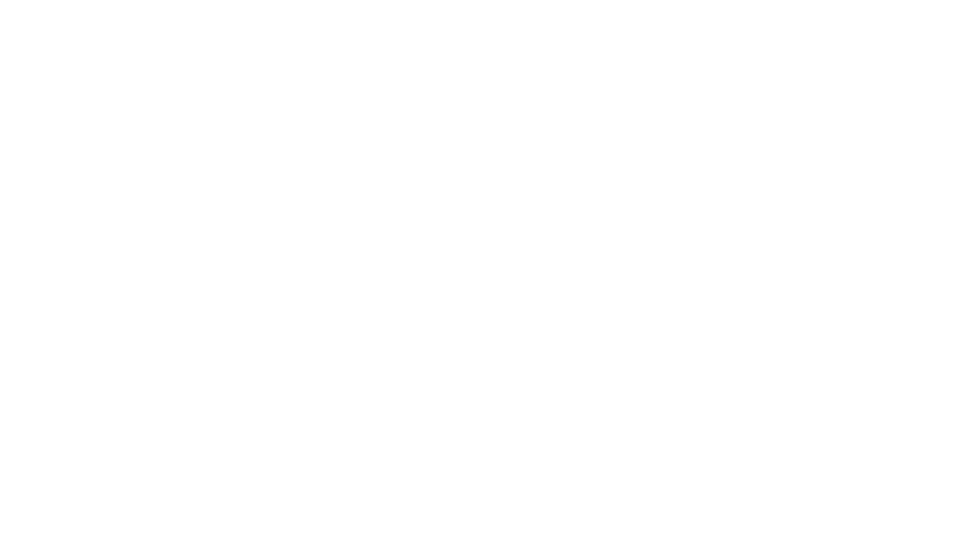}
    \caption{\textbf{Qualitative Results on PartNet Part Segmentation}. We illustrate our performance compared to the ground truth on PartNet. }
    \label{fig:qualitative_partnet_appendix}
\end{figure*}

 Figures \ref{fig:qualitative_dilation}, \ref{fig:qualitative_nearest_neighbors}, \ref{fig:qualitative_layers}, \ref{fig:qualitative_filters}, \ref{fig:qualitative_56_28W} show qualitative results for DeepGCNs on S3DIS \cite{2017arXiv170201105A} and \figLabel \ref{fig:qualitative_partnet_appendix} shows qualitative results for DeepGCNs on PartNet \cite{Mo_2019_CVPR}.

\section{Run-time Overhead of Dynamic k-NN} \label{appendix:runtime_overhead}
We conduct a run-time experiment comparing the inference time of the reference model \emph{ResGCN-28} (28 layers, $k$=16) with dynamic k-NN and fixed k-NN. The inference time with fixed k-NN is 45.63ms. Computing the dynamic k-NN increases the inference time by 150.88ms. It is possible to reduce computation by updating the k-NN less frequently (\eg computing the dynamic k-NN every 3 layers).

\section{Comparison with DGCNN over All Classes} \label{appendix:dgcnn_all_classes}
To showcase the consistent improvement of our framework over the baseline DGCNN \cite{wang2018dynamic}, we reproduce the results of DGCNN\footnote{The results across all classes were not provided in the DGCNN paper.} in \tblLabel \ref{tbl:ours_vs_dgcnn} and find our method outperforms DGCNN in all classes.

\begin{table}[h]
\small
\centering
\setlength{\tabcolsep}{6pt} 
\begin{tabular}{l|cc}
\toprule
\textbf{Class}  & \textbf{DGCNN \cite{wang2018dynamic}} & \textbf{ResGCN-28 (\textit{Ours})} \\
\midrule
{ceiling} & 92.7 & \textbf{93.1} \\
{floor}   & 93.6 & \textbf{95.3} \\
{wall}    & 77.5 & \textbf{78.2} \\
{beam}    & 32.0 & \textbf{33.9} \\
{column}  & 36.3 & \textbf{37.4} \\
{window}  & 52.5 & \textbf{56.1} \\
{door}    & 63.7 & \textbf{68.2} \\
{table}   & 61.1 & \textbf{64.9} \\
{chair}   & 60.2 & \textbf{61.0} \\
{sofa}    & 20.5 & \textbf{34.6} \\
{bookcase}& 47.7 & \textbf{51.5} \\
{board}   & 42.7 & \textbf{51.1} \\
{clutter} & 51.5 & \textbf{54.4} \\
\midrule
\textbf{mIOU}    & 56.3 & \textbf{60.0} \\
\bottomrule
\end{tabular}
\vspace{2pt}
\caption{\textbf{Comparison of \emph{ResGCN-28} with DGCNN}. Average per-class results across all areas for our reference network with 28 layers, \emph{residual graph connections} and \emph{dilated graph convolutions} compared to DGCNN baseline. \emph{ResGCN-28} outperforms DGCNN across all the classes. Metric shown is IoU.}
\label{tbl:ours_vs_dgcnn}
\end{table}

\end{document}